\documentclass[sigconf]{acmart}

\usepackage{colortbl}
\usepackage{bbding}
\usepackage{tabularx}
\usepackage[normalem]{ulem}
\usepackage{amsmath,amsfonts}
\usepackage{algorithmic}
\usepackage{graphicx}
\usepackage{textcomp}
\usepackage{multicol}
\usepackage{caption}
\usepackage{xspace}
\usepackage{multirow}
\usepackage{booktabs}
\usepackage{bm}
\usepackage{xcolor}
\usepackage{makecell}
\usepackage{subcaption}
\usepackage{enumitem}
\usepackage[most]{tcolorbox}
\usepackage{minted}
\tcbuselibrary{minted, breakable}

\newcommand{\cmark}{\ding{51}}
\newcommand{\xmark}{\ding{55}}

\usepackage{pifont}
\AtBeginDocument{%
  }

\setcopyright{acmlicensed}
\copyrightyear{2018}
\acmYear{2018}
\acmDOI{https://doi.org/10.1145/3770855.3819049}
\acmConference[Conference acronym 'XX]{Make sure to enter the correct
  conference title from your rights confirmation email}{June 03--05,
  2018}{Woodstock, NY}
\acmISBN{978-1-4503-XXXX-X/2018/06}

\acmSubmissionID{852}
\begin{document}

\title{Rewarding the Scientific Process: Process-Level Reward Modeling for Agentic Data Analysis}

\author{Zhisong Qiu}
\affiliation{%
  \institution{Zhejiang University}
  \institution{Ant Group}
  \institution{Zhejiang University - Ant Group Joint Laboratory of Knowledge Graph}
  \city{Hangzhou}
  \country{China}
}
\email{qiuzhisong@zju.edu.cn}

\author{Shuofei Qiao}
\author{Kewei Xu}
\affiliation{%
  \institution{Zhejiang University}
  \city{Hangzhou}
  \country{China}
}
\email{kewe1x@zju.edu.cn}

\author{Yuqi Zhu}
\affiliation{%
  \institution{Zhejiang University}
  \city{Hangzhou}
  \country{China}
}
\email{zhuyuqi@zju.edu.cn}

\author{Lun Du}
\affiliation{%
  \institution{Ant Group}
  \city{Hangzhou}
  \country{China}
}
\email{dulun.dl@antgroup.com}

\author{Ningyu Zhang}
\affiliation{%
  \institution{School of Software Technology, Zhejiang University}
  \city{Hangzhou}
  \country{China}
}
\email{zhangningyu@zju.edu.cn}

\author{Huajun Chen}
\authornote{Corresponding author}
\affiliation{%
  \institution{Zhejiang Key Laboratory of Intelligent Manufacturing for Functional Chemicals, ZJU-Hangzhou Global Scientific and Technological Innovation Center}
  \institution{Zhejiang University}
  \city{Hangzhou}
  \country{China}
}
\email{huajunsir@zju.edu.cn}

\renewcommand{\shortauthors}{Qiu et al.}

\begin{abstract}
Process Reward Models (PRMs) have achieved remarkable success in augmenting the reasoning capabilities of Large Language Models (LLMs) within static domains such as mathematics. However, their potential in dynamic data analysis tasks remains underexplored. In this work, we  first present a empirical study revealing that general-domain PRMs struggle to supervise data analysis agents. Specifically, they fail to detect silent errors, logical flaws that yield incorrect results without triggering interpreter exceptions, and erroneously penalize exploratory actions, mistaking necessary trial-and-error exploration for grounding failures. To bridge this gap, we introduce DataPRM, a novel environment-aware generative process reward model that (1) can serve as an active verifier, autonomously interacting with the environment to probe intermediate execution states and uncover silent errors, and (2) employs a reflection-aware ternary reward strategy that distinguishes between correctable grounding errors and irrecoverable mistakes. We design a scalable pipeline to construct over 7K high-quality training instances for DataPRM via diversity-driven trajectory generation and knowledge-augmented step-level annotation. Experimental results demonstrate that DataPRM improves downstream policy LLMs by 7.21\% on ScienceAgentBench and 11.28\% on DABStep using Best-of-N inference. Notably, with only 4B parameters, DataPRM outperforms strong baselines, and exhibits robust generalizability across diverse Test-Time Scaling strategies. Furthermore, integrating DataPRM into Reinforcement Learning yields substantial gains over outcome-reward baselines, achieving 78.73\% on DABench and 64.84\% on TableBench, validating the effectiveness of process reward supervision\footnote{Code: \url{https://github.com/zjunlp/DataMind}.}.
\end{abstract}





\maketitle

\begin{figure*}[t]
    \centering
    \includegraphics[width=0.95\linewidth]{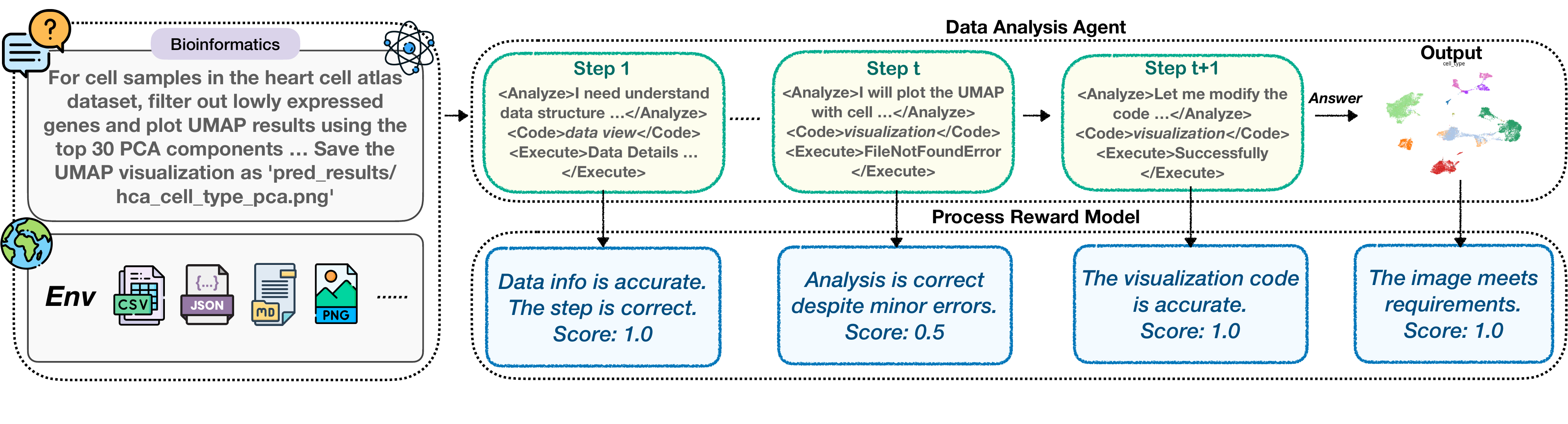}
    % \vskip -6mm
    \caption{The Collaborative Pipeline Between Data Analysis Agent and Process Reward Model (PRM). 
    The agent addresses data analysis tasks while the PRM supervises the agent's procedural steps.}
    \label{fig:teaser_comparison}
\end{figure*}

\section{Introduction}
\label{sec:intro}

Automated data science, aiming to autonomously generate novel scientific knowledge or hypotheses from complex datasets, stands as a core objective in modern scientific discovery \citep{dr_survey, dsagent_survey}.
Central to this pursuit is automated data analysis, the key step to derive evidence-based insights and supportive scientific conclusions to help human decision-making.
As Large Language Models (LLMs) have demonstrated remarkable reasoning capabilities \citep{reasoning-survey,long-cot-survey} on a wide spectrum of tasks such as mathematics \citep{deepseekmathv2, deepmindmath, deepseek_prover, seed-prover, math_survey} and science \citep{ai-scientist, ai4research, x-master, agent_laboratory, sci_claim_lora}, researchers are now increasingly locating them as the backbone of data analysis agents to automate the scientific discovery pipeline \citep{datainterpreter, datamind, deepanalyze, agenticdata, dsstar, agentada, datawiseagent, datacopilot, dagent}.
%However, prevailing approaches only focus on outcome supervision, overlooking the intricate, multi-step nature of data analysis tasks.
%This outcome-centric paradigm entails significant risks in scientific contexts, as it can frequently propagate hallucinated logic, potentially leading to plausible yet scientifically invalid discoveries.
However, prevailing approaches focus only on outcome supervision, overlooking the multi-step rigor of data analysis. In scientific research, where the process must be error-free, this outcome-centric paradigm risks propagating hallucinated logic, yielding seemingly plausible but invalid discoveries.

Conversely, Process Reward Models (PRMs) have exhibited remarkable success in domains such as mathematical reasoning \citep{omegaprm, qwenprm, math-shepherd, genprm, reasonflux-prm, thinkprm} and code generation \citep{codeprm, orps, funprm}.
By providing step-level supervision and fine-grained verification during both training and inference time, PRMs can significantly boost the models' reasoning reliability and performance boundary \citep{1btts, tts, prmsurvey, verengine}.
Despite their proven efficacy, the application of step-level supervision in the domain of data analysis remains largely unexplored.
This leads to a key question: \textbf{\textit{How can we effectively implement step-level supervision for automated data analysis tasks?}}

To bridge this gap, we first analyze the cross-domain applicability of state-of-the-art general PRMs to data-analytic tasks.
Our preliminary analysis reveals that existing PRMs fail to reliably verify two specific categories of errors inherent to this domain: (1) \textit{Silent Errors}: General PRMs struggle to identify logical flaws that yield incorrect results without triggering interpreter exceptions. (2) \textit{Grounding Errors}: they often mistake necessary trial-and-error exploration for irrecoverable failures, leading to early penalization. 
These findings indicate that off-the-shelf PRMs are insufficient for reliable process supervision in data analysis.

Driven by these insights, we introduce \textbf{DataPRM}, a novel Process Reward Model tailored specifically for data analysis agents. Unlike previous PRMs designed for static reasoning tasks, DataPRM can interact dynamically with the environment to validate steps based on real-world data contexts, thereby avoiding deception by mere code execution success.
Furthermore, DataPRM employs a ternary reward strategy to distinguish between incorrect steps, correct steps, and neutral exploratory steps, preventing the suppression of necessary exploration.
To construct DataPRM, we design a scalable data generation pipeline utilizing diversity-driven trajectory generation and knowledge-augmented expert annotation, yielding over 7K high-quality supervision instances.
We apply DataPRM in both Test-Time Scaling (TTS) and Reinforcement Learning (RL) frameworks to further boost the performance boundary of current data analysis agents.

We evaluate DataPRM across multiple data analysis benchmarks.
In TTS settings, incorporating a 4B-parameter DataPRM improves downstream policy models by 7.21\% on ScienceAgentBench and 11.28\% on DABStep. Notably, our model outperforms powerful baselines, such as self-rewarding strategies using Qwen3-235B-A22B-Instruct, while achieving 58$\times$ parameter efficiency. In RL settings, models trained with our process supervision achieve 78.73\% on DABench and 64.84\% on TableBench, surpassing methods relying solely on outcome supervision. Our extensive analysis offers two valuable insights to the community: (1) Environment interaction is critical for process supervision in data analysis; (2) In scenarios with vast exploration spaces, the diversity of supervision steps may outweigh the strict specialization of annotations.
DataPRM not only improves LLM-based data analysis reliability but also provides a scalable framework for fine-grained process supervision in scientific discovery.

In summary, the main contributions of this work are as follows:
\begin{itemize}[leftmargin=*]
    \item We propose DataPRM, a novel process reward model that utilizes environment interaction and ternary rewards to resolve the grounding and silent error challenges in automated data analysis.
    \item We introduce a robust pipeline for generating fine-grained process supervision data, producing a dataset of over 7K annotated instances through diversity-driven trajectory generation and knowledge-augmented step-level annotation.
    \item We empirically validate DataPRM in both TTS and RL settings, achieving significant performance gains on benchmarks such as ScienceAgentBench, while demonstrating $58\times$ parameter efficiency over comparable large-scale baselines.
\end{itemize}
\begin{table*}
\renewcommand\arraystretch{1.}
\caption{\textbf{Representative cases for both error types. Key errors are highlighted in red alongside their descriptions and the PRM's misjudged reward, illustrating the blind spots of current PRMs.}}
\centering
\scalebox{0.9}{
\begin{tabular}{l|p{0.2\linewidth}|p{0.3\linewidth}|p{0.3\linewidth}}
\toprule
\textbf{Category} &
\textbf{Error Explanation} &
\textbf{Wrong Agent Step} &
\textbf{Error Description \& PRM output} \\
\midrule
\multirow{1}{*}{Silent Error} &
\parbox{1.0\linewidth}{
The code interpreter executes successfully. \newline
But the actual result obtained is incorrect.} &
\parbox{1.0\linewidth}{
Step 5: <Analyze>Now, we will create a visualization of proposed roads on the deforestation: ... \colorbox{gray!20}{5.5 km risky buffer} \colorbox{gray!20}{(in light coral/pink)}... We'll save the plot to `pred\_results/predictedRiskyArea.png'. </Analyze> \newline <Code>\textit{visualization code}</Code> \newline
<Execute>\colorbox{gray!20}{Successfully saved}</Execute>}
& \parbox{1.0\linewidth}{
The data analysis agent claims to have drawn the 5.5 km risky buffer (in light coral/pink) and saved the visualization image, \colorbox{gray!20}{but the buffer is not actually successfully} \colorbox{gray!20}{drawn in the visualization}.

\textbf{PRM $\rightarrow$ \{correct\}} \xmark \ \textcolor{red}{Fail to detect the silent error}} \\

\midrule
\multirow{1}{*}{Grounding Error} &
\parbox{1.0\linewidth}{The model's prior knowledge conflicts with actual data  in the environment.} &
\parbox{1.0\linewidth}{
Step 1: <Analyze>The goal is to compare R² results between single-task and multi-task models across different datasets from the TDC ADMET benchmark. I'll start by loading the data and examining its structure.</Analyze> \newline
<Code>\textit{data view code}</Code> \newline
<Execute>\colorbox{gray!20}{KeyError: `dataset'}</Execute>}
& \parbox{1.0\linewidth}{
The data analysis agent encounters a KeyError error during the initial attempt to load the file. \colorbox{gray!20}{The actual key is 'Dataset'}. This is part of the agent's environment exploration.

\textbf{PRM $\rightarrow$ \{incorrect\}} \xmark \ \textcolor{red}{Detect the error but not realize that the agent is in environment exploration.}} \\
\bottomrule
\end{tabular}}
\label{tab:errors}
\end{table*}
\begin{figure*}[t] 
    \centering
    \begin{subfigure}[b]{0.33\textwidth} 
        \centering
        \includegraphics[width=\textwidth]{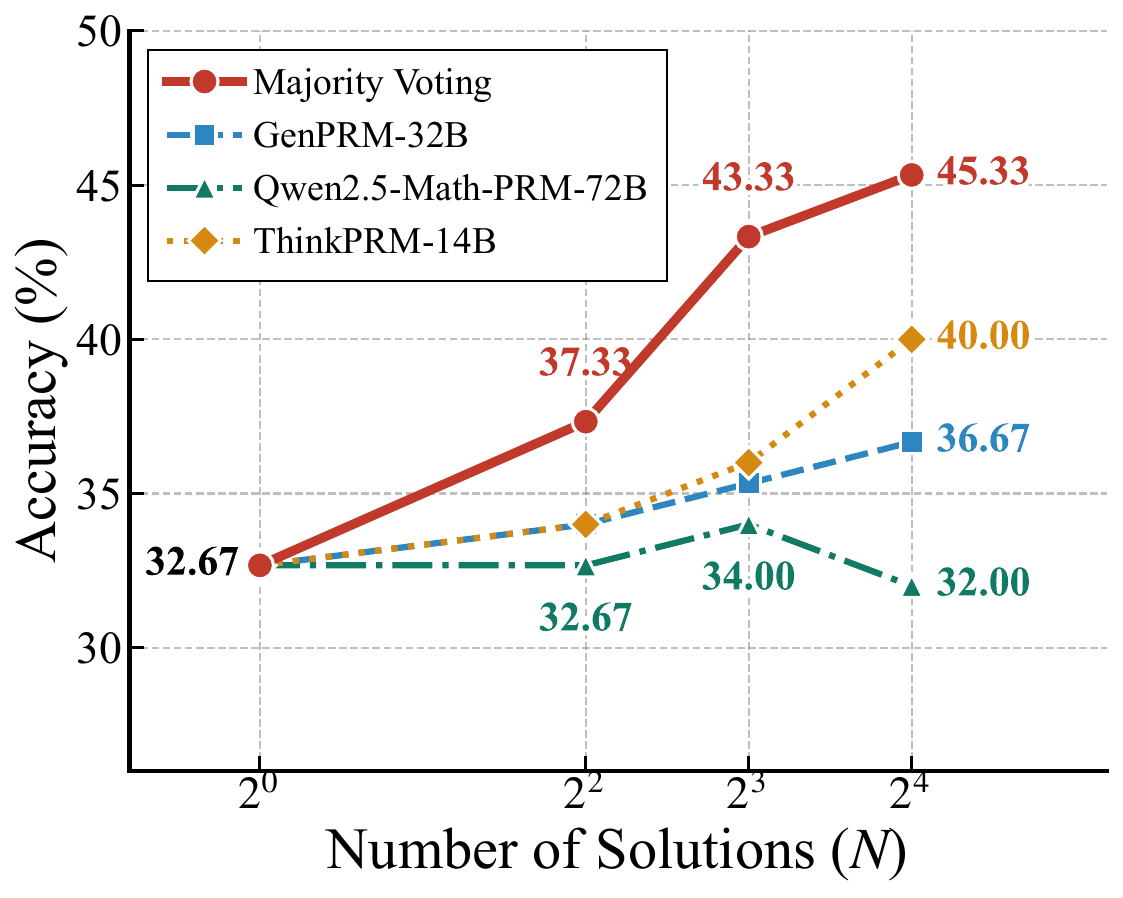}
        \caption{General PRMs' Performance.} 
        \label{fig:math_prm}
    \end{subfigure}
    \begin{subfigure}[b]{0.33\textwidth} 
        \centering
        \includegraphics[width=1.0\textwidth]{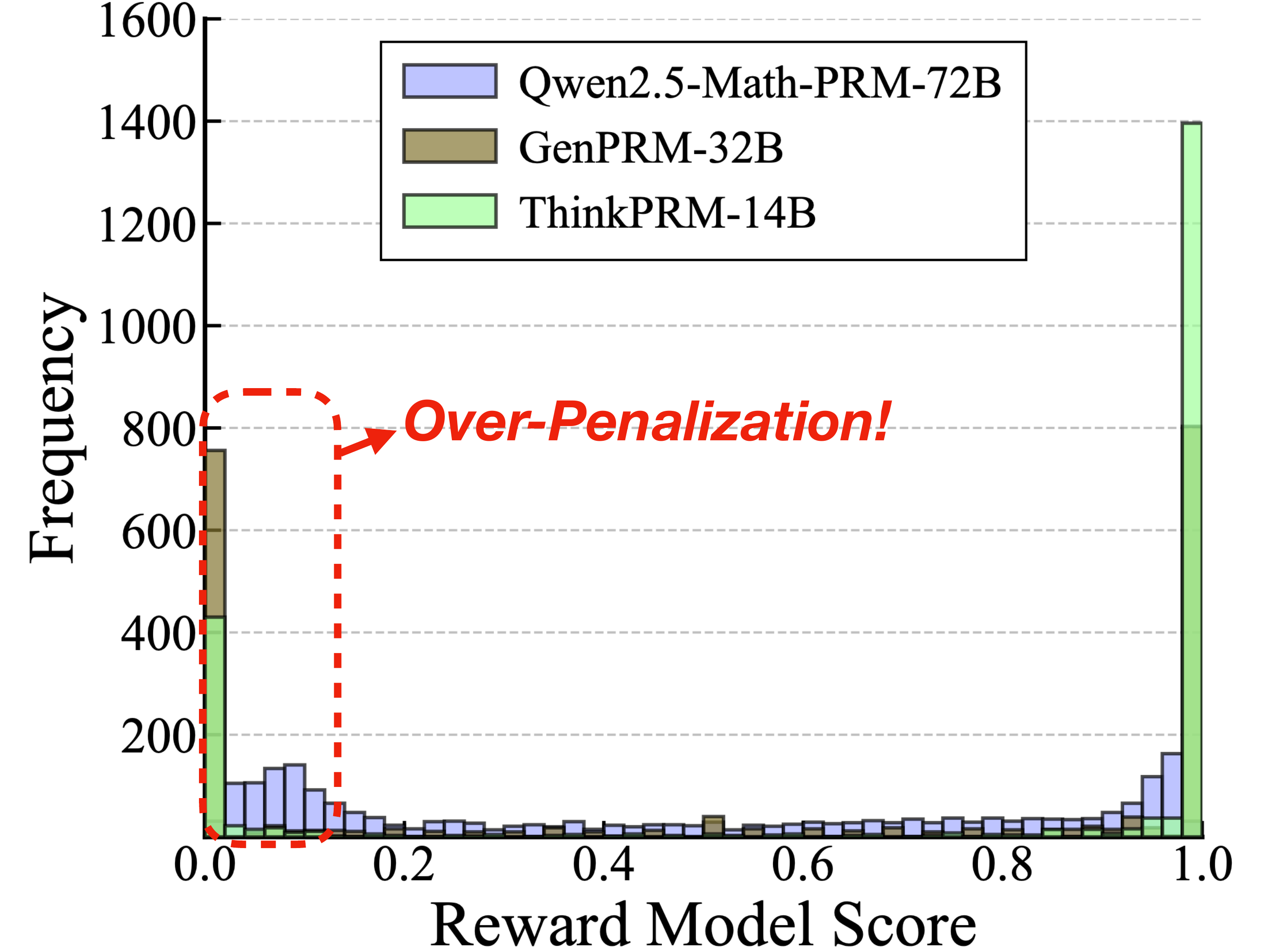}
        \caption{Score Distribution for Grounding Errors.} 
        \label{fig:reflection}
    \end{subfigure}
    \begin{subfigure}[b]{0.33\textwidth}
        \centering
        \includegraphics[width=\textwidth]{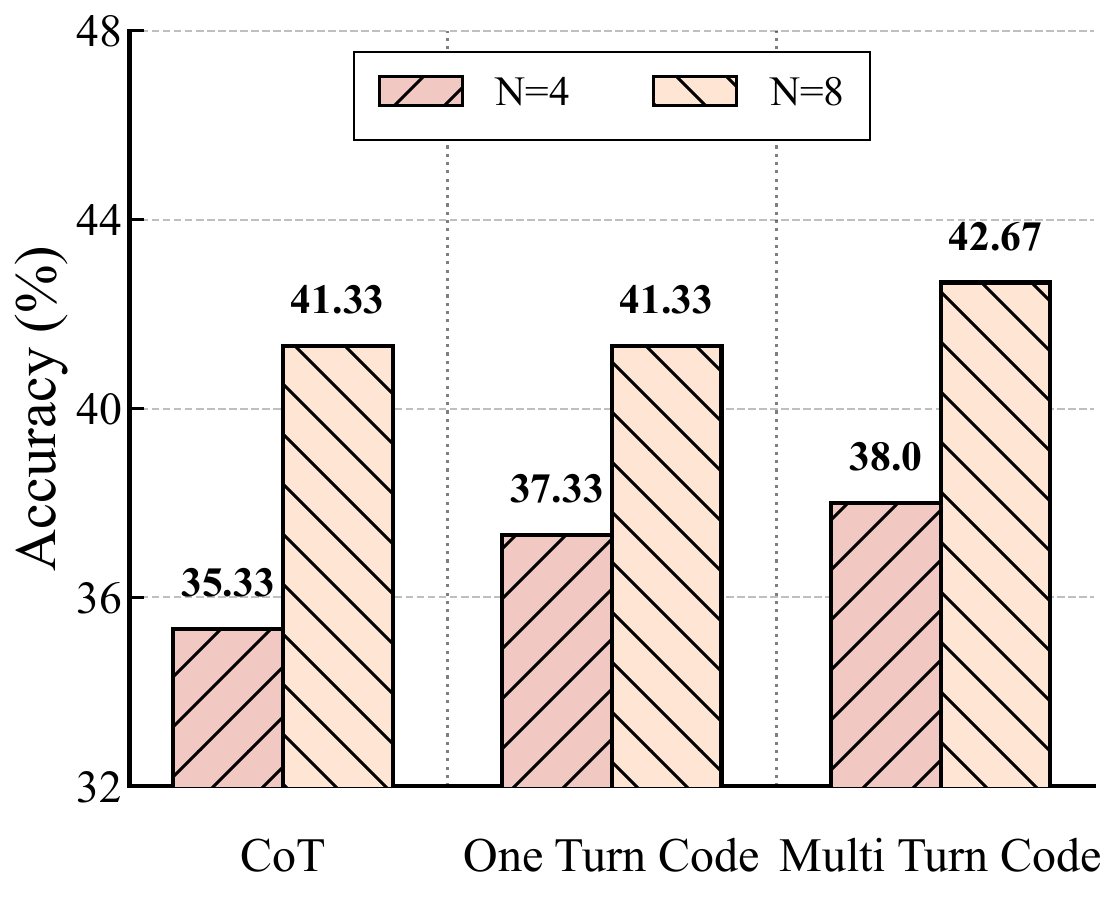}
        \caption{Ablation on Environment Interaction.}
        \label{fig:prompt_prm}
    \end{subfigure}

    \caption{(a): General PRMs' Best-of-N performance on a subset of DABStep. (b): General PRMs' scores on steps with grounding errors despite correct final answers. (c): Ablation study on environment interaction based on prompted LLMs.}
    
\end{figure*}

\section{Preliminary}
\subsection{Data-Analytic Agents}
We formalize the data analysis process as a Partially Observable Markov Decision Process, denoted by the tuple $(\mathcal{U}, \mathcal{S}, \mathcal{A}, \mathcal{T}, \mathcal{O})$. Here, the state space $\mathcal{S}$ characterizes the environment, which typically comprises a code interpreter $\mathcal{I}$ and a set of files $\mathcal{F}$. The process commences with a specific task $u \in \mathcal{U}$ associated with an initial environmental state $s_0 \in \mathcal{S}$. Given the current state $s$, the agent performs an action $a \in \mathcal{A}$ through code generation.
The code interpreter $\mathcal{I}$ also functions as the state transition mechanism, $T(s'|s, a) \in \mathcal{T}$, determining the subsequent state $s'$. Under the assumption of partial observability, the agent perceives the current state solely through an observation $o \in \mathcal{O}$ from the interpreter. Then the historical interaction trajectory at time $t$ can be represented as $h_t = (u, a_0, o_0, a_1, o_1, \dots, a_{t-1}, o_{t-1})$. In scenarios adopting the \texttt{ReAct} \cite{react} framework, where explicit reasoning $z$ guides action generation, the trajectory can be finally formulated as:
\begin{align}
    h_t = (u, z_0, a_0, o_0, z_1, a_1, o_1, \dots, z_{t-1}, a_{t-1}, o_{t-1}).
\end{align}
In our problem setup, the components $z_t, a_t, o_t$ at time step $t$ are regarded as a unified step $\tau_t$ of data analytic agents.

\subsection{Reward Modeling for Data Analysis}
As illustrated in Fig.\ref{fig:teaser_comparison}, given a data-analytic agent's historical interaction trajectory $h_t$ and the current step $\tau_t$, a standard Process Reward Model (PRM) parameterized by $\theta$, utilizes a scoring function $R_\theta(\cdot)$ to assign a step-level reward $r_t$.
The overall trajectory-level reward $r_{traj}$ is then derived by aggregating these step-level rewards:
\begin{align}
    r_t \sim R_\theta( \cdot |h_t, \tau_t), \text{with}\ r_{traj}=\mathcal{A}(r_1, r_2, \dots, r_T)
\end{align}
where $\mathcal{A}(\cdot)$ represents an aggregation function, typically \textsc{Sum} or \textsc{Mean} \citep{1btts}. By providing either step-level reward $r_t$ or trajectory-level reward $r_{traj}$, the verifier can not only enhance the policy model's reasoning performance through search algorithms (e.g. Best-of-N or Beam Search), but also provide more fine-grained reward signals for reinforcement learning.
\section{General PRMs on Data Analysis Tasks}
\label{sec:motivation}

We begin by assessing the efficacy of existing general-domain PRMs in supervising data analysis agents.
Specifically, we conduct a pilot study to investigate two Research Questions (RQs):

\begin{figure*}[t]
    \centering
    \includegraphics[width=0.9\textwidth]{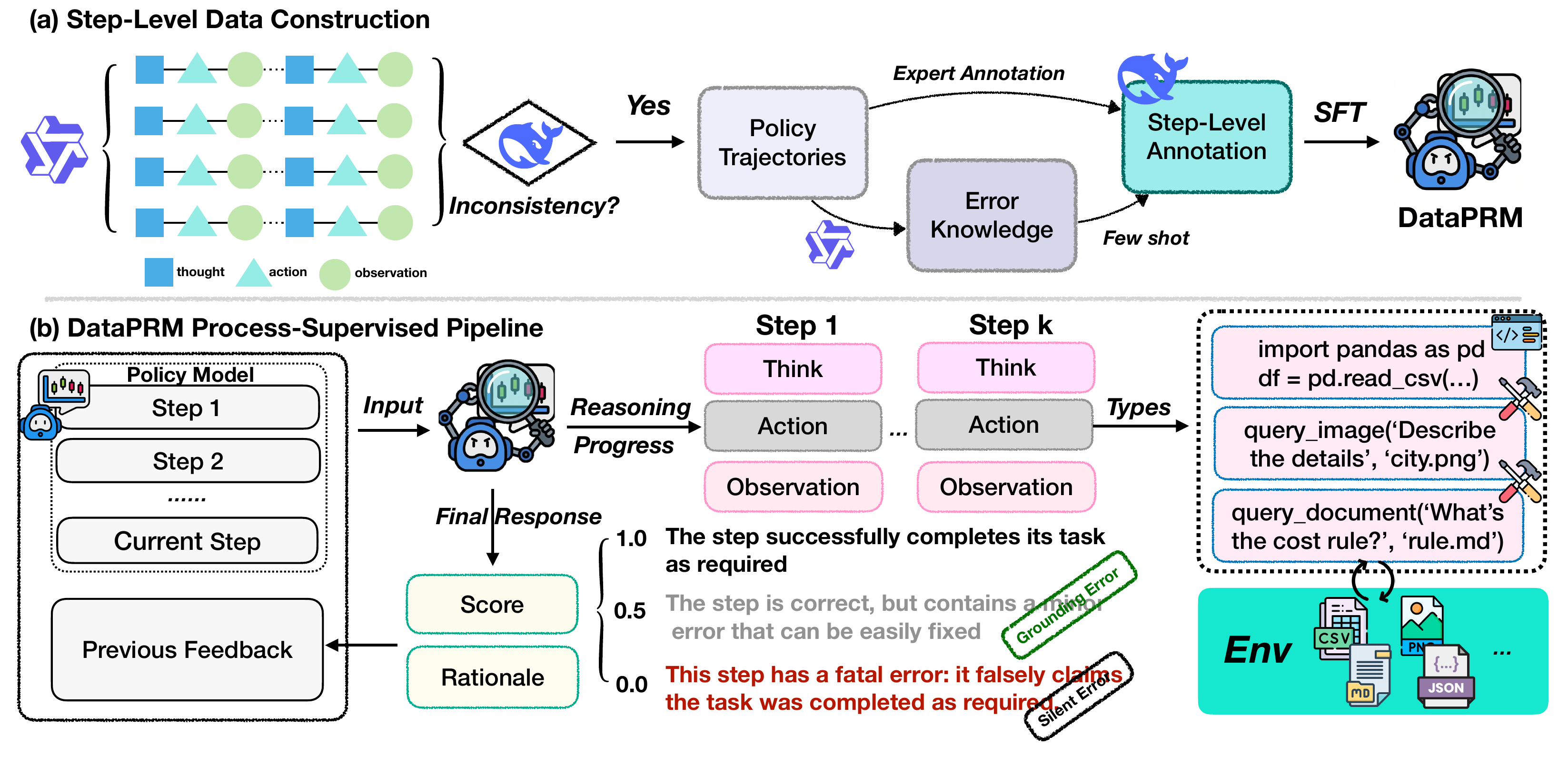}
    \caption{Overview of DataPRM Framework. (a): A diversity-driven trajectory generation strategy followed by knowledge-augmented step-level annotation. (b): DataPRM employs multi-turn interaction, tool-augmented capability and reflection-aware reward strategy for scoring.}
    \label{fig:method_frame}
\end{figure*}

\begin{tcolorbox}[
    enhanced,
    colback=yellow!10!white,      %
    colframe=black,            %
    coltitle=white,            %
    fonttitle=\bfseries,       %
    boxrule=.7pt,
    width=\linewidth,
    top=1mm,
    bottom=1mm,
    left=2mm,                  %
    right=2mm,                 %
    before skip=6pt, after skip=6pt,
    attach boxed title to top left={
        yshift=-2mm,
        xshift=2mm
    },
    boxed title style={
        colback=black,
        sharp corners,
        boxrule=0pt,
        top=2pt, bottom=2pt, left=4pt, right=4pt
    }，
]
\textit{\textbf{RQ1 -}} Can general-domain PRMs effectively distinguish valid reasoning steps in data analysis tasks compared to simple ensemble baselines (e.g., Majority Voting)?
\end{tcolorbox}
% \vspace{3pt}

\begin{tcolorbox}[
    enhanced,
    colback=yellow!10!white,      %
    colframe=black,            %
    coltitle=white,            %
    fonttitle=\bfseries,       %
    boxrule=.7pt,
    width=\linewidth,
    top=1mm,
    bottom=1mm,
    left=2mm,                  %
    right=2mm,                 %
    before skip=6pt, after skip=6pt,
    attach boxed title to top left={
        yshift=-2mm,
        xshift=2mm
    },
    boxed title style={
        colback=black,
        sharp corners,
        boxrule=0pt,
        top=2pt, bottom=2pt, left=4pt, right=4pt
    }，
]
\textit{\textbf{RQ2 -}} What are the specific failure modes of current PRMs when supervising data analysis agents, particularly regarding environment interaction?
\end{tcolorbox}

We utilize Qwen3-235B-A22B-Instruct \citep{qwen3} as the policy model and evaluate on a subset of the DABStep benchmark.

\paragraph{Performance Bottleneck of General PRMs} To address RQ1, we benchmark three state-of-the-art PRMs (Qwen2.5-Math-PRM-72B \citep{qwenprm}, GenPRM \citep{genprm}, and ThinkPRM \citep{thinkprm}) against a Majority Voting baseline. As shown in Fig.\ref{fig:math_prm}, while PRM-guided search (Best-of-N) improves over single-path generation (e.g., ThinkPRM improves performance from 32.67\% to 40.00\% at N=16), it surprisingly fails to surpass the Majority Voting baseline. This suggests that general-domain PRMs lack the specific discriminative capability required for data analysis, rendering them less cost-effective than simple sampling strategies. 

\paragraph{Error Analysis: Grounding vs. Silent Errors} To answer RQ2, we perform a fine-grained error analysis and identify two critical failure modes (Tab.\ref{tab:errors}) that baffle current PRMs:

\textbf{Misjudgment of Exploratory Failures (Grounding Errors).} Data analysis agents often encounter ``Grounding Errors'' —syntax or schema errors arising from a lack of prior knowledge about the data file (e.g., guessing a wrong column name). These are often recoverable and necessary steps for the agent to learn the environment through feedback. We collect steps that contain grounding errors but yield correct final answers, and have the existing PRMs score them. As Fig.\ref{fig:reflection} shows, existing PRMs often treat these steps as fatal errors, assigning them low scores. This penalizes exploration and causes the search algorithm to prune trajectories that would have led to a correct solution after self-correction.

\begin{tcolorbox}[
    enhanced,
    colback=blue!5!white,      %
    colframe=black,            %
    coltitle=white,            %
    fonttitle=\bfseries,       %
    boxrule=0.5pt,
    width=\linewidth,
    top=1mm,
    bottom=1mm,
    left=2mm,                  %
    right=2mm,                 %
    before skip=6pt, after skip=6pt,
    attach boxed title to top left={
        yshift=-2mm,
        xshift=2mm
    },
    boxed title style={
        colback=black,
        sharp corners,
        boxrule=0pt,
        top=2pt, bottom=2pt, left=4pt, right=4pt
    }
]
\textbf{Takeaway 1:} Existing PRMs fail to distinguish between fatal errors and recoverable exploratory steps, often penalizing the latter harshly and impeding environment adaptation.
\end{tcolorbox}

\textbf{Inability to Detect Silent Errors.} Conversely, ``Silent Errors'' occur when code executes without exceptions but produces incorrect results due to logical flaws. Since current PRMs rely primarily on static reasoning (reading the code text), they cannot verify the semantic correctness of the execution result. As shown in Fig.\ref{fig:prompt_prm}, we employ in-context learning to have Qwen3-30B-A3B-Instruct \cite{qwen3} function as the PRM and evaluate it on the same subset of DABStep. We observe that when the PRM is granted the ability to actively interact with the environment (via one-turn code or multi-turn code), it can more accurately select the correct steps. Moreover, the performance under the multi-turn setting surpasses that of the one-turn setting. This is likely because, in the multi-turn setting, the PRM can attempt more interaction to verify the correctness.

\begin{tcolorbox}[
    enhanced,
    colback=blue!5!white,      %
    colframe=black,            %
    coltitle=white,            %
    fonttitle=\bfseries,       %
    boxrule=0.5pt,
    width=\linewidth,
    top=1mm,
    bottom=1mm,
    left=2mm,                  %
    right=2mm,                 %
    before skip=6pt, after skip=6pt,
    attach boxed title to top left={
        yshift=-2mm,
        xshift=2mm
    },
    boxed title style={
        colback=black,
        sharp corners,
        boxrule=0pt,
        top=2pt, bottom=2pt, left=4pt, right=4pt
    }
]
\textbf{Takeaway 2:} PRMs with environment interaction capability can better verify the correctness of data analysis steps.
\end{tcolorbox}

\paragraph{Motivation for DataPRM} Our analysis reveals that the core limitation of current methods is the lack of an environment-aware verifier. We need a PRM that can (1) forgive recoverable grounding errors to encourage exploration, and (2) actively interact with the data to catch silent errors. Motivated by these observations, we introduce a novel process reward model specifically tailored to enhance data-analytic agents.

\section{Methodology}

\subsection{Environment-Aware Verifier Architecture}
We introduce \textbf{DataPRM}, an environment-aware generative PRM. It adopts the \texttt{ReAct} paradigm and can interact with the environment.

\subsubsection{Generative ReAct Paradigm for Verification}
We argue that effective verification in data analysis requires as many contextual interaction capabilities as the solution generation itself. Consequently, our PRM is modeled using the same \texttt{ReAct} paradigm as the data analysis agent.

Given a trajectory $h_t$ of policy model and its current step $\tau_t$ at time $t$, the input context for DataPRM is:
\begin{equation}\label{eq:input}
h_{t,0}^{prm} = h_t \oplus \tau_t = h_t \oplus (z_t, a_t, o_t)
\end{equation}
where $\oplus$ denotes sequence concatenation. This ensures the reward model judges the current step $a_t$ in light of the entire problem-solving trajectory $h_t$ and its immediate outcome $o_t$.
Then DataPRM engages in a multi-step reasoning and verification process. Let $k$ denote the internal time step. At each step $k$, the DataPRM generates a verification tuple $\kappa_{t,k} = (\hat{z}_k, \hat{a}_k, \hat{o}_k)$.
Then the internal context updates as follows:
\begin{equation}
h_{t,k+1}^{prm} = h_{t,k}^{prm} \oplus \kappa_{t,k}
\end{equation}
This internal \texttt{ReAct} loop continues until the DataPRM decides to terminate at step $K$. The final action $\hat{a}_K$ is no longer in code form, but rather a verification result composed of a score and a rationale. Let $\rho_{\phi}$ denote the DataPRM, the final output is:
\begin{equation}
(\hat{z}_{K}, r_t, c_t)\sim \rho_{\phi}(\cdot | h_{t,K}^{prm})
\end{equation}
Here, $r_t$ is the scalar quality score for the step $\tau_t$ of the policy model, and $c_t$ is the explanatory rationale derived from the verification trajectory. And the feedback tuple $(r_t, c_t)$ generated by the DataPRM is not discarded but is explicitly appended to the context for the next time step $t+1$ verification. Given the historical verification result $f_{t}=(r_0,c_0,r_1,c_1,\dots,r_{t-1},c_{t-1})$ of the verifier, we redefine the input of DataPRM in Formula \ref{eq:input} as follows:
\begin{equation}
h_{t,0}^{prm} = h_t \oplus f_{t} \oplus \tau_t
\end{equation}
This form ensures that DataPRM can access verification information from previous steps, thereby guaranteeing the consistency and continuity of the evaluation. Additionally, we provide a theoretical perspective in Appx.\ref{app:theoretical_perspective}.

\subsubsection{Tool-Augmented Capability Integration}
When interacting with a data analysis environment, PRMs may require multiple capabilities, such as multimodal understanding (reading images) or long-context comprehension (reading manual documents). Recent studies indicate that LLM agents can autonomously leverage tools to engage with external environments and progressively refine their reasoning ability \citep{retool, toolrl, tattoo, autogen}. Inspired by their work, we decouple the verifier's capabilities into \textit{intrinsic reasoning} (acquired via training) and \textit{extrinsic perception} (acquired via tools).
We equip DataPRM with two tools, namely \texttt{query\_document} and \texttt{query\_image}. DataPRM can query related questions about documents or images through function calls in the code, and the tools will invoke the corresponding expert models to provide answers.
By bridging internal code generation with external tool usage, DataPRM achieves comprehensive verification coverage across data files, manual documents, and images.

\subsubsection{Reflection-Aware Reward Strategy}
\label{sec:reflection-reward-strategy}
As shown in \S\ref{sec:motivation}, existing PRMs cannot distinguish between grounding errors and other types when assigning scores.
To address this, we expand the step-level reward space $r_t \in \{0, 1\}$ to a ternary set $\mathcal{R} = \{0, 0.5, 1\}$ to capture the nuance of agentic behaviors:

\begin{itemize}[leftmargin=*]
    \item \textbf{Strictly Correct ($r_t=1.0$):} The step is logically sound and it advances the solution directly.
    \item \textbf{Irrecoverable Error ($r_t=0.0$):} The step contains fundamental logic flaws or hallucinations that steer the trajectory to a dead end from which recovery is impossible.
    \item \textbf{Correctable Error ($r_t=0.5$):} The step contains a minor error (e.g. syntax error, incorrect file path) but effectively triggers an environment feedback loop that allows for potential correction.
\end{itemize}

\subsection{Step-Level Data Construction}
Existing public data analysis datasets rarely provide both source files and fine-grained step annotations, making off-the-shelf process-supervised training difficult. We therefore introduce a data generation pipeline with diversity-driven trajectory generation and knowledge-augmented step-level annotation.

\subsubsection{Diversity-Driven Trajectory Generation}

We primarily adapt the AutoSDT \citep{autosdt} methodology to crawl GitHub for files related to scientific data analysis.
To increase the volume of usable data, human experts revise and extend a subset of these files.
For query generation, we use DeepSeek-V3.2 \citep{deepseek} as an expert model to synthesize reasoning-focused queries, while directly adopting validated AutoSDT queries for visualization tasks.
For each validated query $x$ from the collection phase, we employ Qwen3-235B-A22B-Instruct \citep{qwen3} as the policy model $\pi_\theta$ to perform parallel sampling.
We generate $K=4$ distinct trajectories and use a judge model $\mathcal{M}$ based on DeepSeek-V3.2 \citep{deepseek} to determine whether their final answers are mutually inconsistent.
To maximize the information gain during PRM training, we retain the trajectory set $\{y^i\}_{i=1}^K$ only when the final answers are not all identical.
This strategy focuses the data on informative boundary cases.

\subsubsection{Knowledge-Augmented Step-Level Annotation.}
We convert collected trajectories into discrete steps.
Qwen3-235B-A22B-Instruct \citep{qwen3} conducts an initial pass for step annotation and error attribution.
To systematically categorize failures, we apply the AutoManual \citep{automanual} framework to merge similar error categories.
Human experts manually verify the rationales of these merged categories and inject them as structured few-shot examples into the annotation prompt.
DeepSeek-V3.2 then assigns final step-level rewards using the ternary reward strategy defined in Section \ref{sec:reflection-reward-strategy}, constructing the final dataset for process supervision.
For quality control, we filter non-analytical errors, such as timeouts and broken files, and verify LLM annotations against human experts before scaling up the pipeline.
Based on 100 manual spot checks, the model achieves 86.0\% raw accuracy and a quadratic weighted Cohen's $\kappa$ of 0.83, confirming high reliability.
\definecolor{mygrey}{RGB}{213, 213, 213}
\newcommand{\GG}{\cellcolor{mygrey}}

\begin{table*}
\renewcommand\arraystretch{1.1}
\caption{\textbf{Main results on ScienceAgentBench and DABStep.} We compare DataPRM against various step verifiers using best-of-$N$ sampling ($N \in \{4, 8, 16\}$) with Qwen3-235B-A22B-Instruct-2507 as the base policy. Best results are in \textbf{bold}. DataPRM achieves state-of-the-art TTS performance using substantially fewer parameters.}
\centering
\scalebox{0.9}{
\begin{tabular}{l|c|ccc|ccc|ccc|ccc}
\toprule
\multirow{3}{*}{\textbf{Verifier (Best-of-N)}} & 
\multirow{3}{*}{\textbf{Params}} & 
\multicolumn{3}{c|}{\textbf{ScienceAgentBench}} & 
\multicolumn{9}{c}{\textbf{DABStep}} \\ 
\cmidrule(l{0.1em}r{0.1em}){3-5} \cmidrule(l{0.1em}){6-14}
& & 
\multicolumn{3}{c|}{\textbf{SR}} & 
\multicolumn{3}{c|}{\textbf{Easy}} & 
\multicolumn{3}{c|}{\textbf{Hard}} & 
\multicolumn{3}{c}{\textbf{Avg.}} \\
\cmidrule(l{0.1em}r{0.1em}){3-5} \cmidrule(l{0.1em}r{0.1em}){6-8} \cmidrule(l{0.1em}r{0.1em}){9-11} \cmidrule(l{0.1em}r{0.1em}){12-14}
& & 4 & 8 & 16 & 4 & 8 & 16 & 4 & 8 & 16 & 4 & 8 & 16 \\
\midrule
Majority Vote & - & \textbf{24.36} & 24.36 & 23.08 & \textbf{75.00} & \textbf{76.39} & 76.39 & 26.98 & 29.63 & 30.69 & 34.66 & 37.11 & 38.00 \\
LLM-as-a-judge & - & \textbf{24.36} & 24.36 & 24.36 & \textbf{75.00} & \textbf{76.39} & 75.00 & 25.13 & 27.51 & 29.63 & 33.11 & 35.33 & 36.89 \\
Self-Rewarding & - & \textbf{24.36} & 24.36 & 24.36 & \textbf{75.00} & \textbf{76.39} & 76.39 & 28.04 & 30.16 & 32.80 & 35.55 & 37.56 & 39.77 \\
Math-Shepherd-PRM-7B & 7B & 19.23 & 21.79 & 20.51 & \textbf{75.00} & 75.00 & 75.00 & 23.28 & 23.28 & 19.31 & 31.56 & 31.56 & 28.22 \\
Qwen2.5-Math-PRM-7B & 7B & 19.23 & 20.51 & 19.23 & \textbf{75.00} & 72.22 & 73.61 & 20.90 & 18.25 & 14.55 & 29.56 & 26.89 & 24.00 \\
ReasonFlux-PRM-7B & 7B & 19.23 & 21.79 & 19.23 & 73.61 & 72.22 & 75.00 & 20.63 & 17.99 & 13.76 & 29.11 & 26.67 & 23.56 \\
ThinkPRM & 14B & 19.23 & 21.79 & 17.95 & \textbf{75.00} & 75.00 & 72.22 & 25.13 & 24.34 & 26.72 & 33.11 & 32.45 & 34.00 \\
GenPRM & 32B & 21.79 & 20.51 & 20.51 & \textbf{75.00} & 73.61 & 73.61 & 24.60 & 25.40 & 26.72 & 32.66 & 33.11 & 34.22  \\
Qwen2.5-Math-PRM-72B & 72B & 23.08 & 23.08 & 20.51 & 73.61 & \textbf{76.39} & 75.00 & 21.96 & 22.75 & 20.37 & 30.22 & 31.33 & 29.11 \\
\midrule
\textbf{DataPRM} & \textbf{4B} & \textbf{24.36} & \textbf{25.64} & \textbf{25.64} & \textbf{75.00} & \textbf{76.39} & \textbf{77.78} & \textbf{29.89} & \textbf{32.80} & \textbf{33.86} & \textbf{37.11} & \textbf{39.77} & \textbf{40.89} \\
\bottomrule
\end{tabular}
}
\label{tab:bon_results}
\end{table*}
\subsection{End-to-End RL Training with PRM}

For end-to-end RL training, we employ the Group Relative Policy Optimization (GRPO) \citep{grpo} algorithm with several effective strategies such as clip-higher and token-level loss to ensure stable optimization \citep{dapo}.
Define $\varrho_{i,t}(\theta) = \frac{\pi_\theta(o_t|q, o_{<t})}{\pi_{\theta_\text{old}}(o_t|q, o_{<t})}$, the loss is:

\begin{equation}\label{eq:dapo}
\begin{aligned}
    &\mathcal{J}(\theta) = \mathbb{E}_{(q, a)\sim \mathcal{D},\{o_i\}_{i=1}^{G}\sim\pi_{\theta_\text{old}}(\cdot|q)} \\
    &\;\;\frac{1}{\sum_{i=1}^{G}|o_i|}\sum_{i=1}^{G}\sum_{t=1}^{|o_i|}\left\{\min \left[\varrho_{i,t}(\theta)\hat{A}_{i,t}, \text{clip}(\varrho_{i,t}(\theta), 1-\epsilon_{l}, 1+\epsilon_{h})\hat{A}_{i,t}\right]\right\}
\end{aligned}
\end{equation}

The total reward $r_{\text{total}}$ is formulated as a weighted combination of the outcome reward $r_{\text{outcome}}$ and the PRM scores $r_{\text{prm}}$:
\begin{equation}
r_{\text{total}} = (1-\beta) \cdot r_{\text{outcome}} + \beta \cdot (\frac{1}{T}\sum_{t=1}^T r_{\text{prm}}(\tau_t))
\end{equation}
where $\beta$ controls the trade-off between outcome correctness and process validity and $\tau_t$ is the step of agent at time $t$. With a group size of $G$, we calculate the group-normalized advantage $\hat{A}_{i,t}$ for the $i$-th output as:
\begin{equation}
\hat{A}_{i,t} = \frac{r_{\text{total}, i} - \text{mean}(\{r_{\text{total}, j}\}^G_{j=1})}{\text{std}(\{r_{\text{total}, j}\}^G_{j=1})}
\end{equation}

Moreover, we observe that discrepancies may arise between the ground truth outcome and the PRM's final step estimation. To address this, we enforce a consistency check:
\begin{equation}
    r_{\text{prm}}(\tau_T) \leftarrow 
    \begin{cases} 
        r_{\text{outcome}} & \text{if } r_{\text{prm}}(\tau_T) \neq r_{\text{outcome}} \\
        r_{\text{prm}}(\tau_T) & \text{otherwise}
    \end{cases}
\end{equation}
This alignment ensures that the model does not learn from conflicting signals at the termination of a trajectory.

\begin{figure}[t]
    \centering
    \includegraphics[width=\linewidth]{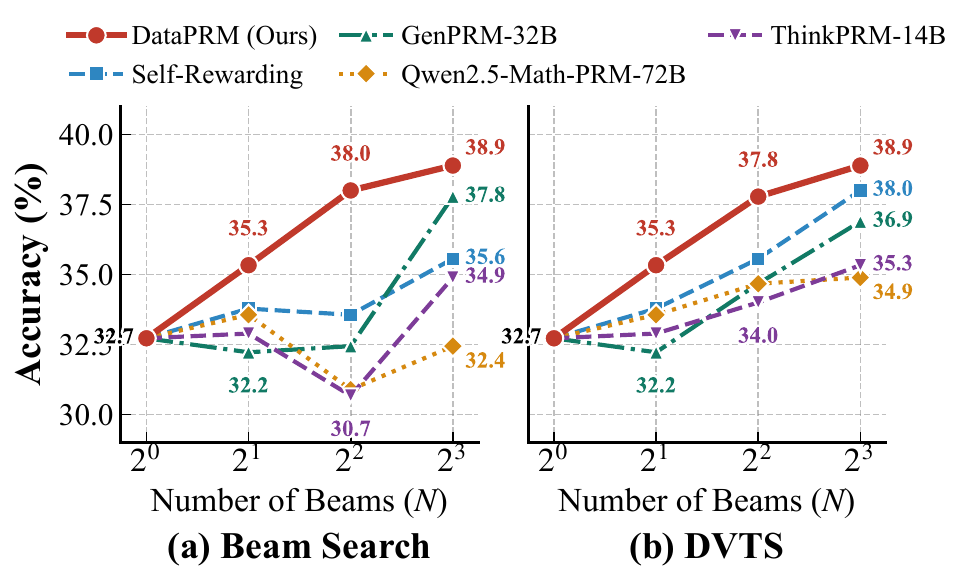}
    \caption{Performance of DataPRM evaluated under two extended TTS strategies, namely: (a) Beam Search and (b) Diverse Verifier Tree Search (DVTS).}
    \label{fig:search_policy}
\end{figure}
\section{Experiments}
\subsection{Experiment Settings}
We first empirically evaluate DataPRM on the Test-Time Scaling (TTS) experiment (Section \ref{sec:main_results}), then conduct an in-depth analysis (Section \ref{sec:analyses}), and finally, we apply DataPRM to Reinforcement Learning (RL) and perform experiments (Section \ref{sec:apply_rl}).

\begin{table*}[htbp]
\centering
\caption{Ablation study of different components. "Env", "Multi", and "Refl" denote the Code Environment, Multi-turn Interaction, and Reflection-aware/Ternary Reward Strategy, respectively. Best results are in \textbf{bold}.}
\label{tab:ablation}
\scalebox{1.0}{
\begin{tabular}{l ccc ccc ccc ccc}
\toprule
\multirow{2}{*}{\textbf{Variant}} & \multicolumn{3}{c}{\textbf{Components}} & \multicolumn{3}{c}{\textbf{Easy}} & \multicolumn{3}{c}{\textbf{Hard}} & \multicolumn{3}{c}{\textbf{Avg.}} \\
\cmidrule(lr){2-4} \cmidrule(lr){5-7} \cmidrule(lr){8-10} \cmidrule(lr){11-13}
 & Env & Multi & Refl & 4 & 8 & 16 & 4 & 8 & 16 & 4 & 8 & 16 \\
\midrule
CoT & \xmark & \xmark & \xmark & 75.00 & 75.00 & 75.00 & 25.93 & 29.37 & 32.01 & 33.78 & 36.67 & 38.89 \\
Single-turn Code w/ Env& \cmark & \xmark & \xmark & \textbf{76.39} & \textbf{76.39} & 76.39 & 28.57 & 30.95 & 32.80 & 36.22 & 38.22 & 39.77 \\
Multi-turn Code w/o Env & \xmark & \cmark & \xmark & 73.61 & 75.00 & 76.39 & 26.46 & 29.89 & 31.75 & 34.00 & 37.11 & 38.89 \\
Multi-turn Code w/ Env & \cmark & \cmark & \xmark & 75.00 & \textbf{76.39} & 76.39 & 29.37 & 30.69 & 32.80 & 36.67 & 38.00 & 39.77 \\
DataPRM & \cmark & \cmark & \cmark & 75.00 & \textbf{76.39} & \textbf{77.78} & \textbf{29.89} & \textbf{32.80} & \textbf{33.86} & \textbf{37.11} & \textbf{39.77} & \textbf{40.89} \\
\bottomrule
\end{tabular}
}
\end{table*}

\subsubsection{Datasets and Metrics}
\label{sec:datasets_metrics}
For TTS, we evaluate DataPRM on two datasets: \textbf{ScienceAgentBench} \citep{scienceagentbench} and \textbf{DABStep} \citep{dabstep}.
For ScienceAgentBench, we filter out ML/DL tasks and retain 78 tasks related to data analysis to ensure a focus on automating data analysis tasks and avoid introducing confounding factors associated with model training processes. And we utilize its provided evaluation procedure to report the Success Rate (SR), in which visualization metrics are assessed by Qwen3-VL-235B-A22B-Instruct \citep{qwen3vl} as a judge. For DABStep, we utilize accuracy as the final evaluation metric. For RL, we evaluate our model on two other datasets: \textbf{DABench} \citep{dabench} and \textbf{TableBench} \citep{tablebench}. We use judge model powered by 
Qwen3-30B-A3B-Instruct \citep{qwen3} to assess the accuracy of the answers, reporting both pass@1 and pass@3 scores.

\subsubsection{Models and Baselines}
\label{sec:models_baselines}
For TTS, we compare DataPRM with various step-level verification baselines, including advanced PRMs, majority voting \citep{1btts}, LLM-as-a-judge \citep{llm_as_judge} using DeepSeek-V3.2 \citep{deepseek}, and self-rewarding \citep{self_rewarding_rm, self_rewarding_prm} utilizing Qwen3-235B-A22B-Instruct \citep{qwen3}. For PRM approaches, we include both discriminative (Qwen-PRM series \citep{qwenprm}, Math-Shepherd-PRM-7B \citep{math-shepherd}, and ReasonFlux-PRM-7B \citep{reasonflux-prm}) and generative (ThinkPRM \citep{thinkprm} and GenPRM \citep{genprm}). 
For the policy reasoning models, we evaluate the proposed method on Qwen3-235B-A22B-Instruct \citep{qwen3}. For RL, we use Qwen2.5-Coder-7B-Instruct as the base model and compare with the SFT model and the model trained with outcome rewards.

\subsubsection{Implementation Details}
\label{sec:models_baselines}
We use ms-swift \citep{ms_swift} for DataPRM SFT training.
For SFT, our learning rate is $1e-5$ with a warmup ratio of $0.05$.
We train 3 epochs and use liger kernel.
Our global batch size is set to $32$.
For DataPRM inference, temperature is $0.7$, the top-p is $0.9$ and the top-k is $20$.
For applying DataPRM to RL, we use verl \citep{verl}.
We use a learning rate of $1e-6$.
The batch size is $32$ with a mini batch size of $2$. The balancing coefficient $\beta$ is set to 0.5.
The rollout temperature is $0.7$, the top-p is $1.0$, and the group size $G$ is $4$. We use \texttt{AgentLoop} and \texttt{RewardLoop} to carry out asynchronous rollout and rewarding, thereby accelerating the training process.
% We train DataPRM on Qwen3-4B-Instruct using ms-swift \citep{ms_swift}. Learning rate is 1e-5, batch size is 32 and training epochs are 3.
All the experiments are conducted on 8 $\times$ H20 GPUs.

\subsection{Main Results}
\label{sec:main_results}
\subsubsection{\textbf{DataPRM Surpasses Larger Baselines with Effective Scaling in Best-of-N}}
Tab.\ref{tab:bon_results} presents the performance of DataPRM and other baselines in the Best-of-N setting.
Although parameterized at only 4B, DataPRM consistently achieves superior results compared to robust baselines like GenPRM-32B and Qwen2.5-Math-PRM-72B. Furthermore, it surpasses both the DeepSeek-V3.2 LLM-as-a-judge framework and the Qwen3-235B-A22B-Instruct self-rewarding baseline. 
Moreover, as $N$ and the number of responses in the candidate pool increase, existing PRMs may discard originally correct responses and select incorrect ones. 
For example, as $N$ expands from 8 to 16, the performance of Qwen2.5-Math-PRM-72B drops from 31.33\% to 29.11\%.  
This indicates that existing PRMs have not truly acquired the ability to distinguish between valid reasoning and hallucinations in data analysis tasks. 
In contrast, DataPRM achieves effective scaling, delivering consistent performance improvements as $N$ increases. 
This suggests that it can discern high-quality data analysis trajectories, thereby providing stronger reward supervision.

\subsubsection{\textbf{DataPRM Generalizes Across Search Strategies and Resists Reward Hacking}}
Beyond best-of-N search, we assess DataPRM under two extended TTS strategies: Beam Search and Diverse Verifier Tree Search (DVTS). These results are then benchmarked against the self-rewarding method and the most competitive PRM baselines.
As shown in Fig.\ref{fig:search_policy}, DataPRM consistently outperforms all baselines across both search strategies and all computation budgets.
Moreover, we observe the instability of other baselines under Beam Search.
For instance, Qwen2.5-Math-PRM-72B exhibits a performance degradation as the search budget increases ($33.56\%\to30.89\%\to32.44\%$).
This phenomenon is often attributed to ``reward hacking'' where the greedy nature of Beam Search exploits inaccuracies in the reward model, leading to high-scoring but incorrect paths.
In contrast, DataPRM maintains a consistent improvement ($35.33\%\to38.00\%\to38.89\%$), indicating the robustness against the exploitative tendencies of search policies.

\begin{table}[t]
\centering
\caption{Ablation study on filtering strategies. Best results are marked in \textbf{bold}.}
\label{tab:filter_ablation}
\begin{tabular}{l ccc}
\toprule
Filter Strategy & 4 & 8 & 16 \\
\midrule
Unfiltered & 37.11 & \textbf{39.77} & \textbf{40.89} \\
Meta-Critic & 36.67 & 36.45 & 40.00 \\
Outcome-Consistency & 36.22 & 38.22 & 39.77 \\
Process-Consistency & \textbf{38.00} & 38.22 & 39.34 \\
\bottomrule
\end{tabular}
\end{table}

\begin{table}[t]
\centering
\caption{Inference cost analysis.}
\label{tab:cost_analysis}
\small
\begin{tabular}{l cccc}
\toprule
Verifier & Total Tokens & Turns & Time(s) & Tool Calls \\ 
\midrule
GenPRM & 7061.25 & 1.00 & 14.86 & - \\ 
Self-Rewarding & 25282.51 & 3.32 & 194.95 & 0.63 \\
DataPRM & 21455.78 & 2.57 & 24.66 & 0.87 \\
DataPRM (parallel) & 21455.78 & 2.57 & 3.30 & 0.87 \\
\bottomrule
\end{tabular}
\end{table}

\begin{figure*}[t]
    \centering
    \begin{subfigure}[b]{0.47\textwidth}
        \centering
        \includegraphics[width=\linewidth]{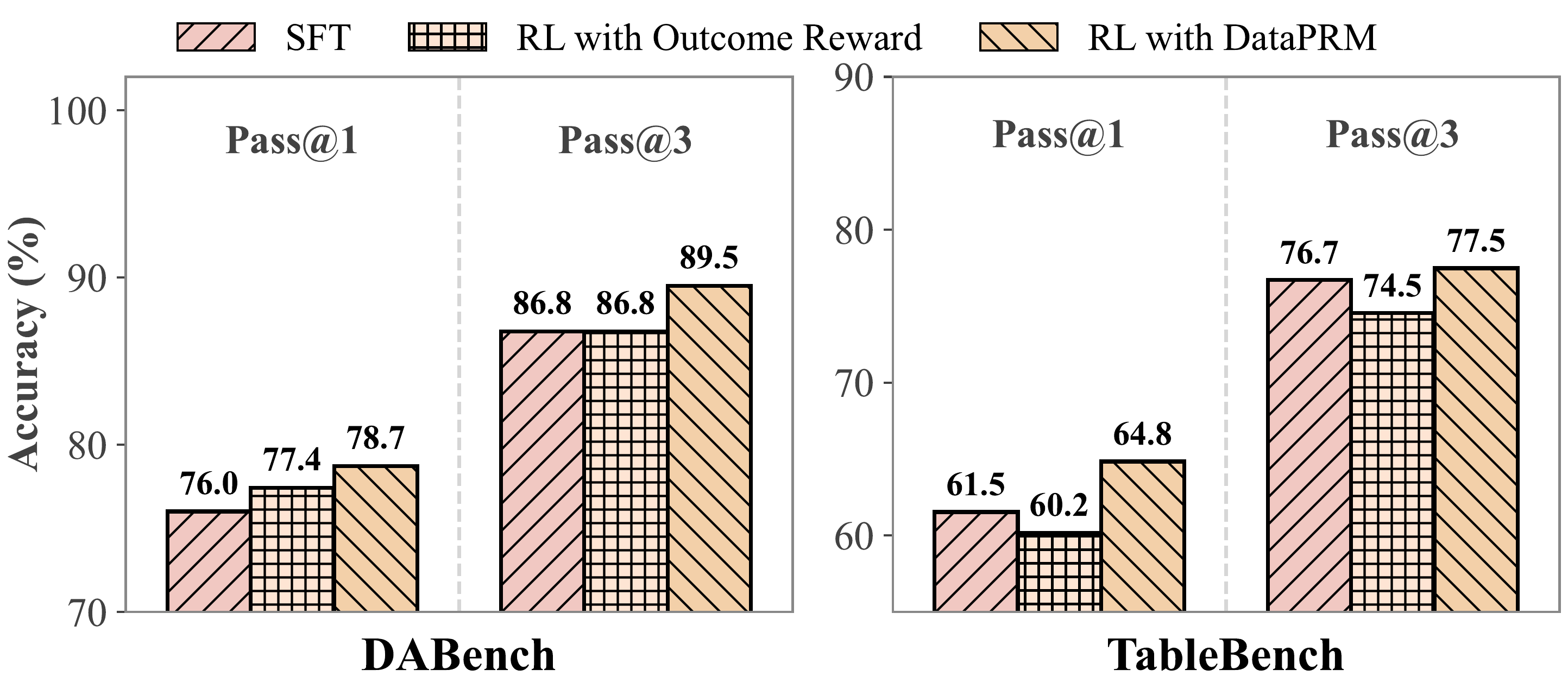}
        \caption{DABench and TableBench Results for RL.}
        \label{fig:rl_merged_results}
    \end{subfigure}
    \hfill 
    \begin{subfigure}[b]{0.245\textwidth}
        \centering
        \includegraphics[width=\linewidth]{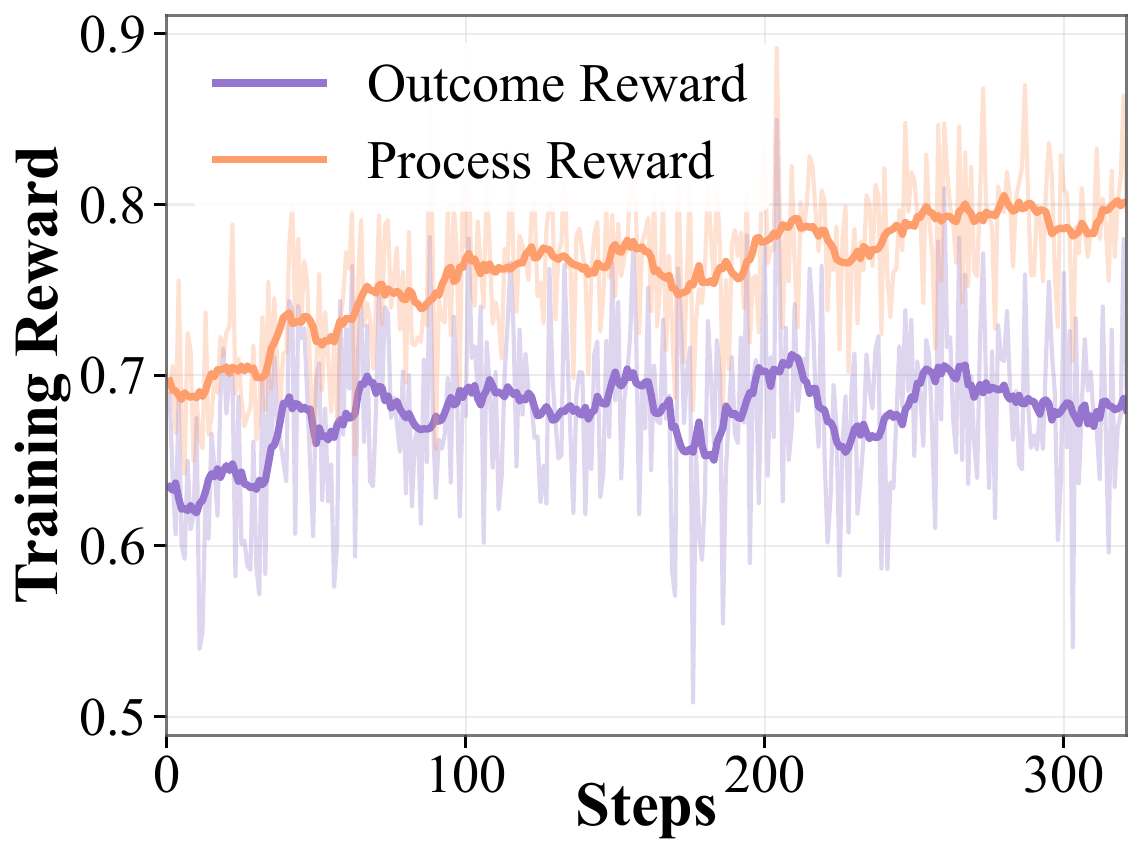}
        \caption{Training Reward Dynamics.}
        \label{fig:rl_reward}
    \end{subfigure}
    \hfill 
    \begin{subfigure}[b]{0.25\textwidth}
        \centering
        \includegraphics[width=\linewidth]{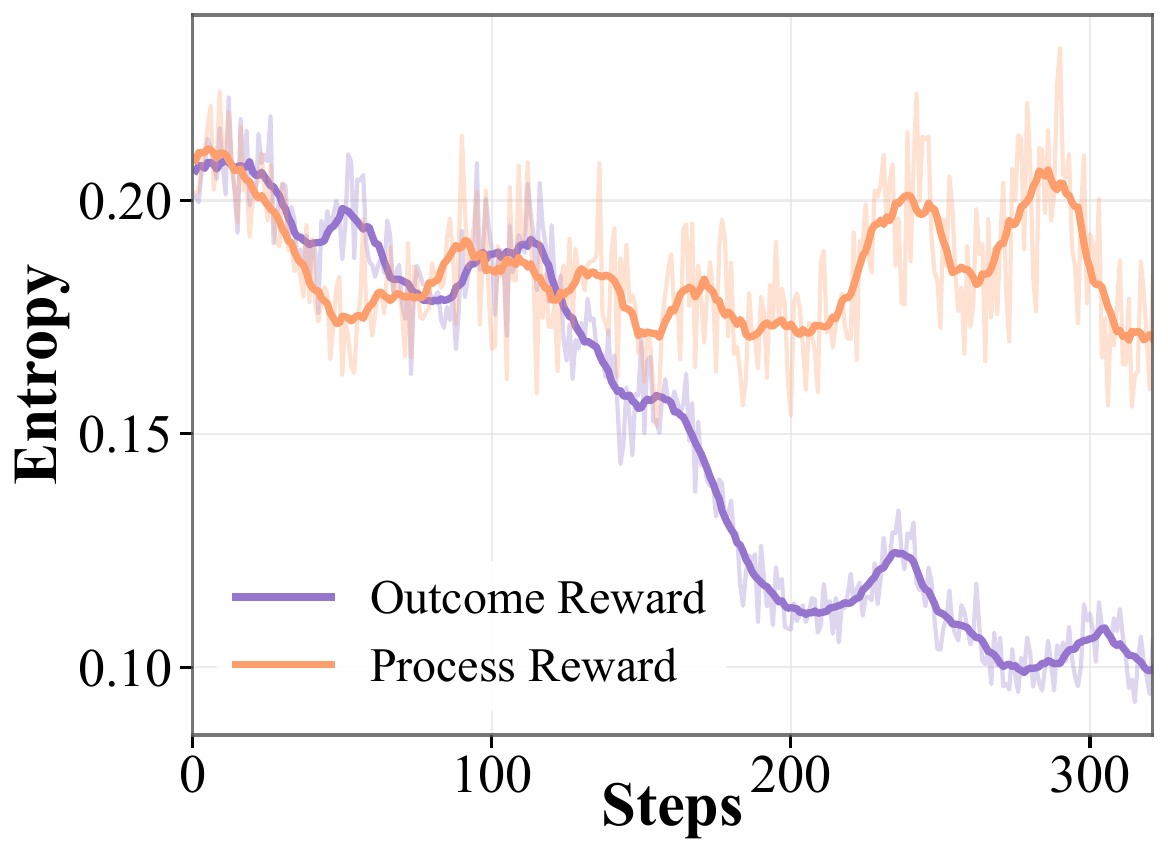}
        \caption{Entropy Dynamics.}
        \label{fig:rl_entropy}
    \end{subfigure}
    \caption{Experiment results on RL training and benchmarks. (a): The evaluation results on DABench and TableBench for models trained with different strategies. (b) and (c): The training reward dynamics and entropy dynamics in RL training for outcome reward and process reward.}
    \label{fig:rl_results}
\end{figure*}

\subsection{In-Depth Analysis}
\label{sec:analyses}
\subsubsection{\textbf{Environment Interaction is Critical for Data Analysis Tasks}}
To assess the efficacy and necessity of individual modules, we conduct an ablation study on the DataPRM architecture.
As shown in Tab.\ref{tab:ablation}, we compared the full method with four variants: (1) CoT (Chain-of-Thought baseline), (2) Single-turn Code w/ Env, (3) Multi-turn Code w/o Env, and (4) Multi-turn Code w/ Env.
First, equipping the model with the ability of environment interaction (Single-turn Code w/ Env) yields a consistent improvement over the CoT baseline, verifying that executable feedback helps ground the reasoning process.
Second, while multi-turn interaction alone provides marginal gains, its combination with the environment (Multi-turn Code w/ Env) significantly boosts performance, suggesting that iterative refinement is most effective when supported by execution results.
After incorporating the reflection-aware strategy, DataPRM achieves optimal performance, demonstrating that assigning fine-grained scores to exploratory steps helps in selecting the correct trajectory.
We also observe that our proposed components are most pronounced on the Hard subset. While the CoT baseline struggles with complex reasoning (32.01\% at $N=16$), introducing the Code Environment and Interaction (Multi-turn Code w/ Env) improves this to 32.80\%. 
DataPRM further raises this to 33.86\%, demonstrating the effectiveness of our method in complex data analysis tasks.

\subsubsection{\textbf{Data Diversity Outweighs Purity for Scalable Reward Modeling}}
\label{sec:filter_study}
Since we do not have any requirement for the correctness of policy model trajectory answers in our data generation process, we explore three types of reference-free trajectory filtering methods \citep{spark}: Meta Critic, Outcome Consistency, and Process Consistency.
Performance comparison with respect to the inference sampling budget $N$ (Best-of-$N$) is reported in Tab.\ref{tab:filter_ablation}.
Counter-intuitively, aggressive filtering does not consistently yield better reward modeling performance.
While Process Consistency achieves a marginal gain at a low sampling budget ($N=4$, 38.00\% vs. 37.11\%), the unfiltered baseline demonstrates superior scalability, significantly outperforming all filtered variants at $N=16$ (40.89\%). We attribute this phenomenon to the trade-off between data purity and diversity.
While strict filtering strategies can enhance data purity, they may also discard other effective and diverse step-wise supervision samples, leading the PRM to become overly conservative.
In contrast, the PRM trained on the full dataset is exposed to the complete trajectory distribution.
By learning from a richer set of step-wise supervision samples, it can more effectively distinguish correct solutions from a larger candidate pool.

\subsubsection{\textbf{DataPRM Provides Efficient Verification for Agentic Workflows}}
\label{sec:cost_analysis}

We further analyze the inference cost of different verifier designs in Tab.~\ref{tab:cost_analysis}.
Compared with GenPRM, DataPRM incurs higher token usage and latency, which is expected because it actively interacts with the execution environment to verify intermediate states.
Specifically, DataPRM uses 2.57 turns and 0.87 tool calls on average, enabling the verifier to ground its judgment in executable feedback rather than relying on textual reasoning.
This additional cost is therefore a necessary trade-off for detecting silent errors and evaluating exploratory data-analysis trajectories.

Despite this overhead, DataPRM is substantially more efficient than the Self-Rewarding baseline.
It reduces total token consumption from 25.3K to 21.5K tokens, corresponding to a 15.1\% reduction, and decreases the average number of turns from 3.32 to 2.57, corresponding to a 22.6\% reduction.
This suggests that DataPRM performs more focused verification by using targeted environment interaction instead of lengthy self-evaluation.
Moreover, we build a parallel evaluation environment with isolated file systems and lightweight string-based context tracking to mitigate the latency overhead.
With this optimization, the practical verification latency is reduced from 24.66s to 3.30s per sample, demonstrating that DataPRM can be deployed in large-scale agentic evaluation without becoming an inference bottleneck.

\subsection{Applying DataPRM to Agentic RL}
\label{sec:apply_rl}
As shown in Fig.\ref{fig:rl_merged_results}, the model trained with process-supervised rewards achieves accuracy rates of 78.73\% on DABench and 64.84\% on TableBench, outperforming both the SFT model and the model trained with outcome-only rewards.
Furthermore, as shown in Fig.\ref{fig:rl_reward} and Fig.\ref{fig:rl_entropy}, a noticeable entropy collapse occurred when training with outcome-only rewards.
After 200 steps, the entropy decreases to approximately 0.12 and the reward ceases to increase.
In contrast, training with the incorporation of process-supervised rewards avoids this phenomenon. The entropy remains around 0.18, and the reward continues to rise steadily.
This indicates that more fine-grained rewards can enable the model to conduct more thorough exploration.
Similarly, the model trained with process-supervised rewards also demonstrates an increase in pass@3, which is likely attributable to the consistently high entropy  maintained throughout training.
In contrast, the model trained with outcome rewards shows no growth in the pass@3 metric.

\section{Related Work}
\subsection{Process Reward Models}
Process Reward Models (PRMs) \cite{PRM800K, prmsurvey, sft_survey} are capable of providing granular rewards and demonstrate significant potential for applications in Test Time Scaling \cite{tts, 1btts, verengine} and Reinforcement Learning \cite{prime, pav, smartsearch, fapo, prm_agentrl}.
Current PRMs primarily focus on scenarios that do not require environmental interaction, such as mathematics \cite{omegaprm, qwenprm, math-shepherd, genprm, reasonflux-prm, thinkprm}, code generation \cite{codeprm, orps, funprm}, tabular reasoning \cite{tattoo, reward-sql, exploretableprm}, and others \citep{toolprmbench, toolprm, finprm}.
Recently, there has been a growing trend of applying PRMs to agent scenarios. Web-Shepherd \cite{web-shepherd} can provide step-wise feedback and reward for web navigation tasks using structured subgoal checklists. AgentPRM \cite{agentprm} employs Temporal Difference-based estimation method combined with Generalized Advantage Estimation, demonstrating excellent performance across multiple agent tasks. SWE-PRM \cite{sweprm} validates that using proprietary models as PRMs can enhance the capabilities of agents in the field of software engineering. To the best of our knowledge, this work represents the first systematic investigation of Process Reward Models (PRMs) within the domain of data analysis, with the broader aim of providing insights for other complex, agent-driven fields.

\subsection{Data-Analytic Agents}
Data analysis agents are aimed at autonomously accomplishing end-to-end data analysis tasks \citep{ddr, scienceagentbench, dabstep, dabench, tablebench, dsgym, dsbench, dataanalysis-study, dataagent_survey}.
To handle complex data analysis problems in real-world scenarios, early approaches primarily relied on prompt engineering and predefined workflows to leverage the reasoning and coding capabilities of closed-source models in addressing these challenges, including data visualization \cite{matplotagent}, insight and report generation \cite{dagent, agentada, insightpilot, datastorm, eff_tool_agent_da}, heterogeneous data analysis \cite{datacopilot, agenticdata, dsstar, datacross}, general data science \cite{datainterpreter, datawiseagent}, etc. Recently, an increasing number of data analysis agents have demonstrated promising performance through agentic training based on open-source models. DataMind \cite{datamind} employs fine-grained query generation, knowledge-based trajectory sampling, and combined agent training paradigm of SFT and RL. DeepAnalyze \cite{deepanalyze} constructs a data-grounded trajectory synthesis framework and employs a curriculum-based agentic training paradigm. Both consistently achieve outstanding performance across multiple data analysis tasks. Unlike methods that rely on predefined workflows or data-driven model training, we utilize PRMs to enhance agents' data analysis capability, offering a novel perspective through the lens of Test Time Scaling.

\section{Conclusion}
In this work, we introduced \textbf{DataPRM}, an environment-aware process reward model designed to overcome the limitations of general PRMs in detecting silent and grounding errors within interactive data analysis. By leveraging active environment verification and a ternary reward strategy, DataPRM provides precise step-level supervision. To construct DataPRM, we designed a scalable data generation pipeline utilizing diversity-driven trajectory generation and knowledge-enhanced expert annotation. Empirical results demonstrate that DataPRM significantly enhances both Test-Time Scaling and Reinforcement Learning performance.

\section{Limitations and Ethical Considerations}
Our current work has several limitations. First, we focus primarily on data analysis tasks involving reasoning and visualization, leaving complex engineering tasks, such as model training and prediction, for future exploration. Second, we train DataPRM solely via Supervised Fine-Tuning (SFT), a paradigm that relies heavily on the availability of high-quality trajectory data. To mitigate this data dependency and further enhance the capabilities of PRMs, our future work will explore methods that require less human-curated data, such as Reinforcement Learning \citep{tattoo} and Skill \citep{skillnet, trace2skill, evoskill, skillx, coevoskills}.

This work follows established ethical research practices, utilizing only synthesized or publicly available datasets. We have accurately cited all sources to ensure transparency and proper attribution.
\begin{acks}
We would like to express sincere gratitude to the  reviewers for their thoughtful and constructive feedback. This work was supported by the National Natural Science Foundation of China (No. 62576307, No. NSFCU23B2055, No. NSFCU19B2027), the Fundamental Research Funds for the Central Universities (226-2023-00138), Yongjiang Talent Introduction Programme (2021A-156-G), and Information Technology Center and State Key Lab of CAD\&CG, Zhejiang University. This work was supported by Ant Group and Zhejiang University - Ant Group Joint Laboratory of Knowledge Graph.
\end{acks}

\bibliographystyle{ACM-Reference-Format}
\bibliography{KDD2026}


\begin{thebibliography}{88}

%%% the Web address.

\ifx \showCODEN    \undefined \def \showCODEN     #1{\unskip}     \fi
\ifx \showISBNx    \undefined \def \showISBNx     #1{\unskip}     \fi
\ifx \showISBNxiii \undefined \def \showISBNxiii  #1{\unskip}     \fi
\ifx \showISSN     \undefined \def \showISSN      #1{\unskip}     \fi
\ifx \showLCCN     \undefined \def \showLCCN      #1{\unskip}     \fi
\ifx \shownote     \undefined \def \shownote      #1{#1}          \fi
\ifx \showarticletitle \undefined \def \showarticletitle #1{#1}   \fi
\ifx \showURL      \undefined \def \showURL       {\relax}        \fi
% invisible to TeX
\providecommand\bibfield[2]{#2}
\providecommand\bibinfo[2]{#2}
\providecommand\natexlab[1]{#1}
\providecommand\showeprint[2][]{arXiv:#2}

\bibitem[Abaskohi et~al\mbox{.}(2025)]%
        {agentada}
\bibfield{author}{\bibinfo{person}{Amirhossein Abaskohi}, \bibinfo{person}{Amrutha~Varshini Ramesh}, \bibinfo{person}{Shailesh Nanisetty}, \bibinfo{person}{Chirag Goel}, \bibinfo{person}{David V{\'{a}}zquez}, \bibinfo{person}{Christopher Pal}, \bibinfo{person}{Spandana Gella}, \bibinfo{person}{Giuseppe Carenini}, {and} \bibinfo{person}{Issam~H. Laradji}.} \bibinfo{year}{2025}\natexlab{}.
\newblock \showarticletitle{AgentAda: Skill-Adaptive Data Analytics for Tailored Insight Discovery}.
\newblock \bibinfo{journal}{\emph{CoRR}}  \bibinfo{volume}{abs/2504.07421} (\bibinfo{year}{2025}).
\newblock
\showeprint[arXiv]{2504.07421}
\href{https://doi.org/10.48550/ARXIV.2504.07421}{doi:\nolinkurl{10.48550/ARXIV.2504.07421}}


\bibitem[Alzubi et~al\mbox{.}(2026)]%
        {evoskill}
\bibfield{author}{\bibinfo{person}{Salaheddin Alzubi}, \bibinfo{person}{Noah Provenzano}, \bibinfo{person}{Jaydon Bingham}, \bibinfo{person}{Weiyuan Chen}, {and} \bibinfo{person}{Tu Vu}.} \bibinfo{year}{2026}\natexlab{}.
\newblock \showarticletitle{EvoSkill: Automated Skill Discovery for Multi-Agent Systems}.
\newblock \bibinfo{journal}{\emph{CoRR}}  \bibinfo{volume}{abs/2603.02766} (\bibinfo{year}{2026}).
\newblock
\showeprint[arXiv]{2603.02766}
\href{https://doi.org/10.48550/ARXIV.2603.02766}{doi:\nolinkurl{10.48550/ARXIV.2603.02766}}


\bibitem[Bai et~al\mbox{.}(2025)]%
        {qwen3vl}
\bibfield{author}{\bibinfo{person}{Shuai Bai}, \bibinfo{person}{Yuxuan Cai}, \bibinfo{person}{Ruizhe Chen}, \bibinfo{person}{Keqin Chen}, \bibinfo{person}{Xionghui Chen}, \bibinfo{person}{Zesen Cheng}, \bibinfo{person}{Lianghao Deng}, \bibinfo{person}{Wei Ding}, \bibinfo{person}{Chang Gao}, \bibinfo{person}{Chunjiang Ge}, \bibinfo{person}{Wenbin Ge}, \bibinfo{person}{Zhifang Guo}, \bibinfo{person}{Qidong Huang}, \bibinfo{person}{Jie Huang}, \bibinfo{person}{Fei Huang}, \bibinfo{person}{Binyuan Hui}, \bibinfo{person}{Shutong Jiang}, \bibinfo{person}{Zhaohai Li}, \bibinfo{person}{Mingsheng Li}, \bibinfo{person}{Mei Li}, \bibinfo{person}{Kaixin Li}, \bibinfo{person}{Zicheng Lin}, \bibinfo{person}{Junyang Lin}, \bibinfo{person}{Xuejing Liu}, \bibinfo{person}{Jiawei Liu}, \bibinfo{person}{Chenglong Liu}, \bibinfo{person}{Yang Liu}, \bibinfo{person}{Dayiheng Liu}, \bibinfo{person}{Shixuan Liu}, \bibinfo{person}{Dunjie Lu}, \bibinfo{person}{Ruilin Luo}, \bibinfo{person}{Chenxu Lv}, \bibinfo{person}{Rui Men},
  \bibinfo{person}{Lingchen Meng}, \bibinfo{person}{Xuancheng Ren}, \bibinfo{person}{Xingzhang Ren}, \bibinfo{person}{Sibo Song}, \bibinfo{person}{Yuchong Sun}, \bibinfo{person}{Jun Tang}, \bibinfo{person}{Jianhong Tu}, \bibinfo{person}{Jianqiang Wan}, \bibinfo{person}{Peng Wang}, \bibinfo{person}{Pengfei Wang}, \bibinfo{person}{Qiuyue Wang}, \bibinfo{person}{Yuxuan Wang}, \bibinfo{person}{Tianbao Xie}, \bibinfo{person}{Yiheng Xu}, \bibinfo{person}{Haiyang Xu}, \bibinfo{person}{Jin Xu}, \bibinfo{person}{Zhibo Yang}, \bibinfo{person}{Mingkun Yang}, \bibinfo{person}{Jianxin Yang}, \bibinfo{person}{An Yang}, \bibinfo{person}{Bowen Yu}, \bibinfo{person}{Fei Zhang}, \bibinfo{person}{Hang Zhang}, \bibinfo{person}{Xi Zhang}, \bibinfo{person}{Bo Zheng}, \bibinfo{person}{Humen Zhong}, \bibinfo{person}{Jingren Zhou}, \bibinfo{person}{Fan Zhou}, \bibinfo{person}{Jing Zhou}, \bibinfo{person}{Yuanzhi Zhu}, {and} \bibinfo{person}{Ke Zhu}.} \bibinfo{year}{2025}\natexlab{}.
\newblock \showarticletitle{Qwen3-VL Technical Report}.
\newblock \bibinfo{journal}{\emph{CoRR}}  \bibinfo{volume}{abs/2511.21631} (\bibinfo{year}{2025}).
\newblock
\showeprint[arXiv]{2511.21631}
\href{https://doi.org/10.48550/ARXIV.2511.21631}{doi:\nolinkurl{10.48550/ARXIV.2511.21631}}


\bibitem[Chae et~al\mbox{.}(2025)]%
        {web-shepherd}
\bibfield{author}{\bibinfo{person}{Hyungjoo Chae}, \bibinfo{person}{Sunghwan Kim}, \bibinfo{person}{Junhee Cho}, \bibinfo{person}{Seungone Kim}, \bibinfo{person}{Seungjun Moon}, \bibinfo{person}{Gyeom Hwangbo}, \bibinfo{person}{Dongha Lim}, \bibinfo{person}{Minjin Kim}, \bibinfo{person}{Yeonjun Hwang}, \bibinfo{person}{Minju Gwak}, \bibinfo{person}{Dongwook Choi}, \bibinfo{person}{Minseok Kang}, \bibinfo{person}{Gwanhoon Im}, \bibinfo{person}{ByeongUng Cho}, \bibinfo{person}{Hyojun Kim}, \bibinfo{person}{Jun~Hee Han}, \bibinfo{person}{Taeyoon Kwon}, \bibinfo{person}{Minju Kim}, \bibinfo{person}{Beong{-}woo Kwak}, \bibinfo{person}{Dongjin Kang}, {and} \bibinfo{person}{Jinyoung Yeo}.} \bibinfo{year}{2025}\natexlab{}.
\newblock \showarticletitle{Web-Shepherd: Advancing PRMs for Reinforcing Web Agents}.
\newblock \bibinfo{journal}{\emph{CoRR}}  \bibinfo{volume}{abs/2505.15277} (\bibinfo{year}{2025}).
\newblock
\showeprint[arXiv]{2505.15277}
\href{https://doi.org/10.48550/ARXIV.2505.15277}{doi:\nolinkurl{10.48550/ARXIV.2505.15277}}


\bibitem[Chai et~al\mbox{.}(2025)]%
        {x-master}
\bibfield{author}{\bibinfo{person}{Jingyi Chai}, \bibinfo{person}{Shuo Tang}, \bibinfo{person}{Rui Ye}, \bibinfo{person}{Yuwen Du}, \bibinfo{person}{Xinyu Zhu}, \bibinfo{person}{Mengcheng Zhou}, \bibinfo{person}{Yanfeng Wang}, \bibinfo{person}{Weinan E}, \bibinfo{person}{Yuzhi Zhang}, \bibinfo{person}{Linfeng Zhang}, {and} \bibinfo{person}{Siheng Chen}.} \bibinfo{year}{2025}\natexlab{}.
\newblock \showarticletitle{SciMaster: Towards General-Purpose Scientific {AI} Agents, Part I. X-Master as Foundation: Can We Lead on Humanity's Last Exam?}
\newblock \bibinfo{journal}{\emph{CoRR}}  \bibinfo{volume}{abs/2507.05241} (\bibinfo{year}{2025}).
\newblock
\showeprint[arXiv]{2507.05241}
\href{https://doi.org/10.48550/ARXIV.2507.05241}{doi:\nolinkurl{10.48550/ARXIV.2507.05241}}


\bibitem[Chen et~al\mbox{.}(2025a)]%
        {seed-prover}
\bibfield{author}{\bibinfo{person}{Jiangjie Chen}, \bibinfo{person}{Wenxiang Chen}, \bibinfo{person}{Jiacheng Du}, \bibinfo{person}{Jinyi Hu}, \bibinfo{person}{Zhicheng Jiang}, \bibinfo{person}{Allan Jie}, \bibinfo{person}{Xiaoran Jin}, \bibinfo{person}{Xing Jin}, \bibinfo{person}{Chenggang Li}, \bibinfo{person}{Wenlei Shi}, \bibinfo{person}{Zhihong Wang}, \bibinfo{person}{Mingxuan Wang}, \bibinfo{person}{Chenrui Wei}, \bibinfo{person}{Shufa Wei}, \bibinfo{person}{Huajian Xin}, \bibinfo{person}{Fan Yang}, \bibinfo{person}{Weihao Gao}, \bibinfo{person}{Zheng Yuan}, \bibinfo{person}{Tianyang Zhan}, \bibinfo{person}{Zeyu Zheng}, \bibinfo{person}{Tianxi Zhou}, {and} \bibinfo{person}{Thomas~Hanwen Zhu}.} \bibinfo{year}{2025}\natexlab{a}.
\newblock \showarticletitle{Seed-Prover 1.5: Mastering Undergraduate-Level Theorem Proving via Learning from Experience}.
\newblock \bibinfo{journal}{\emph{CoRR}}  \bibinfo{volume}{abs/2512.17260} (\bibinfo{year}{2025}).
\newblock
\showeprint[arXiv]{2512.17260}
\href{https://doi.org/10.48550/ARXIV.2512.17260}{doi:\nolinkurl{10.48550/ARXIV.2512.17260}}


\bibitem[Chen et~al\mbox{.}(2024)]%
        {automanual}
\bibfield{author}{\bibinfo{person}{Minghao Chen}, \bibinfo{person}{Yihang Li}, \bibinfo{person}{Yanting Yang}, \bibinfo{person}{Shiyu Yu}, \bibinfo{person}{Binbin Lin}, {and} \bibinfo{person}{Xiaofei He}.} \bibinfo{year}{2024}\natexlab{}.
\newblock \showarticletitle{AutoManual: Constructing Instruction Manuals by {LLM} Agents via Interactive Environmental Learning}. In \bibinfo{booktitle}{\emph{Advances in Neural Information Processing Systems 38: Annual Conference on Neural Information Processing Systems 2024, NeurIPS 2024, Vancouver, BC, Canada, December 10 - 15, 2024}}, \bibfield{editor}{\bibinfo{person}{Amir Globersons}, \bibinfo{person}{Lester Mackey}, \bibinfo{person}{Danielle Belgrave}, \bibinfo{person}{Angela Fan}, \bibinfo{person}{Ulrich Paquet}, \bibinfo{person}{Jakub~M. Tomczak}, {and} \bibinfo{person}{Cheng Zhang}} (Eds.).
\newblock
\urldef\tempurl%
\url{http://papers.nips.cc/paper\_files/paper/2024/hash/0142921fad7ef9192bd87229cdafa9d4-Abstract-Conference.html}
\showURL{%
\tempurl}


\bibitem[Chen et~al\mbox{.}(2025c)]%
        {long-cot-survey}
\bibfield{author}{\bibinfo{person}{Qiguang Chen}, \bibinfo{person}{Libo Qin}, \bibinfo{person}{Jinhao Liu}, \bibinfo{person}{Dengyun Peng}, \bibinfo{person}{Jiannan Guan}, \bibinfo{person}{Peng Wang}, \bibinfo{person}{Mengkang Hu}, \bibinfo{person}{Yuhang Zhou}, \bibinfo{person}{Te Gao}, {and} \bibinfo{person}{Wanxiang Che}.} \bibinfo{year}{2025}\natexlab{c}.
\newblock \showarticletitle{Towards Reasoning Era: {A} Survey of Long Chain-of-Thought for Reasoning Large Language Models}.
\newblock \bibinfo{journal}{\emph{CoRR}}  \bibinfo{volume}{abs/2503.09567} (\bibinfo{year}{2025}).
\newblock
\showeprint[arXiv]{2503.09567}
\href{https://doi.org/10.48550/ARXIV.2503.09567}{doi:\nolinkurl{10.48550/ARXIV.2503.09567}}


\bibitem[Chen et~al\mbox{.}(2025d)]%
        {ai4research}
\bibfield{author}{\bibinfo{person}{Qiguang Chen}, \bibinfo{person}{Ming{-}Hsuan Yang}, \bibinfo{person}{Libo Qin}, \bibinfo{person}{Jinhao Liu}, \bibinfo{person}{Zheng Yan}, \bibinfo{person}{Jiannan Guan}, \bibinfo{person}{Dengyun Peng}, \bibinfo{person}{Yiyan Ji}, \bibinfo{person}{Hanjing Li}, \bibinfo{person}{Mengkang Hu}, \bibinfo{person}{Yimeng Zhang}, \bibinfo{person}{Yihao Liang}, \bibinfo{person}{Yu Zhou}, \bibinfo{person}{Jiaqi Wang}, \bibinfo{person}{Zhi Chen}, {and} \bibinfo{person}{Wanxiang Che}.} \bibinfo{year}{2025}\natexlab{d}.
\newblock \showarticletitle{AI4Research: {A} Survey of Artificial Intelligence for Scientific Research}.
\newblock \bibinfo{journal}{\emph{CoRR}}  \bibinfo{volume}{abs/2507.01903} (\bibinfo{year}{2025}).
\newblock
\showeprint[arXiv]{2507.01903}
\href{https://doi.org/10.48550/ARXIV.2507.01903}{doi:\nolinkurl{10.48550/ARXIV.2507.01903}}


\bibitem[Chen et~al\mbox{.}(2025b)]%
        {scienceagentbench}
\bibfield{author}{\bibinfo{person}{Ziru Chen}, \bibinfo{person}{Shijie Chen}, \bibinfo{person}{Yuting Ning}, \bibinfo{person}{Qianheng Zhang}, \bibinfo{person}{Boshi Wang}, \bibinfo{person}{Botao Yu}, \bibinfo{person}{Yifei Li}, \bibinfo{person}{Zeyi Liao}, \bibinfo{person}{Chen Wei}, \bibinfo{person}{Zitong Lu}, \bibinfo{person}{Vishal Dey}, \bibinfo{person}{Mingyi Xue}, \bibinfo{person}{Frazier~N. Baker}, \bibinfo{person}{Benjamin Burns}, \bibinfo{person}{Daniel Adu{-}Ampratwum}, \bibinfo{person}{Xuhui Huang}, \bibinfo{person}{Xia Ning}, \bibinfo{person}{Song Gao}, \bibinfo{person}{Yu Su}, {and} \bibinfo{person}{Huan Sun}.} \bibinfo{year}{2025}\natexlab{b}.
\newblock \showarticletitle{ScienceAgentBench: Toward Rigorous Assessment of Language Agents for Data-Driven Scientific Discovery}. In \bibinfo{booktitle}{\emph{The Thirteenth International Conference on Learning Representations, {ICLR} 2025, Singapore, April 24-28, 2025}}. \bibinfo{publisher}{OpenReview.net}.
\newblock
\urldef\tempurl%
\url{https://openreview.net/forum?id=6z4YKr0GK6}
\showURL{%
\tempurl}


\bibitem[Cui et~al\mbox{.}(2025)]%
        {prime}
\bibfield{author}{\bibinfo{person}{Ganqu Cui}, \bibinfo{person}{Lifan Yuan}, \bibinfo{person}{Zefan Wang}, \bibinfo{person}{Hanbin Wang}, \bibinfo{person}{Wendi Li}, \bibinfo{person}{Bingxiang He}, \bibinfo{person}{Yuchen Fan}, \bibinfo{person}{Tianyu Yu}, \bibinfo{person}{Qixin Xu}, \bibinfo{person}{Weize Chen}, \bibinfo{person}{Jiarui Yuan}, \bibinfo{person}{Huayu Chen}, \bibinfo{person}{Kaiyan Zhang}, \bibinfo{person}{Xingtai Lv}, \bibinfo{person}{Shuo Wang}, \bibinfo{person}{Yuan Yao}, \bibinfo{person}{Xu Han}, \bibinfo{person}{Hao Peng}, \bibinfo{person}{Yu Cheng}, \bibinfo{person}{Zhiyuan Liu}, \bibinfo{person}{Maosong Sun}, \bibinfo{person}{Bowen Zhou}, {and} \bibinfo{person}{Ning Ding}.} \bibinfo{year}{2025}\natexlab{}.
\newblock \showarticletitle{Process Reinforcement through Implicit Rewards}.
\newblock \bibinfo{journal}{\emph{CoRR}}  \bibinfo{volume}{abs/2502.01456} (\bibinfo{year}{2025}).
\newblock
\showeprint[arXiv]{2502.01456}
\href{https://doi.org/10.48550/ARXIV.2502.01456}{doi:\nolinkurl{10.48550/ARXIV.2502.01456}}


\bibitem[DeepSeek{-}AI(2025)]%
        {deepseek}
\bibfield{author}{\bibinfo{person}{DeepSeek{-}AI}.} \bibinfo{year}{2025}\natexlab{}.
\newblock \showarticletitle{DeepSeek-V3.2: Pushing the Frontier of Open Large Language Models}.
\newblock \bibinfo{journal}{\emph{CoRR}}  \bibinfo{volume}{abs/2512.02556} (\bibinfo{year}{2025}).
\newblock
\showeprint[arXiv]{2512.02556}
\href{https://doi.org/10.48550/ARXIV.2512.02556}{doi:\nolinkurl{10.48550/ARXIV.2512.02556}}


\bibitem[Ding et~al\mbox{.}(2025)]%
        {fapo}
\bibfield{author}{\bibinfo{person}{Yuyang Ding}, \bibinfo{person}{Chi Zhang}, \bibinfo{person}{Juntao Li}, \bibinfo{person}{Haibin Lin}, \bibinfo{person}{Xin Liu}, {and} \bibinfo{person}{Min Zhang}.} \bibinfo{year}{2025}\natexlab{}.
\newblock \showarticletitle{{FAPO:} Flawed-Aware Policy Optimization for Efficient and Reliable Reasoning}.
\newblock \bibinfo{journal}{\emph{CoRR}}  \bibinfo{volume}{abs/2510.22543} (\bibinfo{year}{2025}).
\newblock
\showeprint[arXiv]{2510.22543}
\href{https://doi.org/10.48550/ARXIV.2510.22543}{doi:\nolinkurl{10.48550/ARXIV.2510.22543}}


\bibitem[Egg et~al\mbox{.}(2025)]%
        {dabstep}
\bibfield{author}{\bibinfo{person}{Alex Egg}, \bibinfo{person}{Martin~Iglesias Goyanes}, \bibinfo{person}{Friso Kingma}, \bibinfo{person}{Andreu Mora}, \bibinfo{person}{Leandro von Werra}, {and} \bibinfo{person}{Thomas Wolf}.} \bibinfo{year}{2025}\natexlab{}.
\newblock \showarticletitle{DABstep: Data Agent Benchmark for Multi-step Reasoning}.
\newblock \bibinfo{journal}{\emph{CoRR}}  \bibinfo{volume}{abs/2506.23719} (\bibinfo{year}{2025}).
\newblock
\showeprint[arXiv]{2506.23719}
\href{https://doi.org/10.48550/ARXIV.2506.23719}{doi:\nolinkurl{10.48550/ARXIV.2506.23719}}


\bibitem[Feng et~al\mbox{.}(2025)]%
        {retool}
\bibfield{author}{\bibinfo{person}{Jiazhan Feng}, \bibinfo{person}{Shijue Huang}, \bibinfo{person}{Xingwei Qu}, \bibinfo{person}{Ge Zhang}, \bibinfo{person}{Yujia Qin}, \bibinfo{person}{Baoquan Zhong}, \bibinfo{person}{Chengquan Jiang}, \bibinfo{person}{Jinxin Chi}, {and} \bibinfo{person}{Wanjun Zhong}.} \bibinfo{year}{2025}\natexlab{}.
\newblock \showarticletitle{ReTool: Reinforcement Learning for Strategic Tool Use in LLMs}.
\newblock \bibinfo{journal}{\emph{CoRR}}  \bibinfo{volume}{abs/2504.11536} (\bibinfo{year}{2025}).
\newblock
\showeprint[arXiv]{2504.11536}
\href{https://doi.org/10.48550/ARXIV.2504.11536}{doi:\nolinkurl{10.48550/ARXIV.2504.11536}}


\bibitem[Gandhi et~al\mbox{.}(2025)]%
        {sweprm}
\bibfield{author}{\bibinfo{person}{Shubham Gandhi}, \bibinfo{person}{Jason Tsay}, \bibinfo{person}{Jatin Ganhotra}, \bibinfo{person}{Kiran Kate}, {and} \bibinfo{person}{Yara Rizk}.} \bibinfo{year}{2025}\natexlab{}.
\newblock \showarticletitle{When Agents go Astray: Course-Correcting {SWE} Agents with PRMs}.
\newblock \bibinfo{journal}{\emph{CoRR}}  \bibinfo{volume}{abs/2509.02360} (\bibinfo{year}{2025}).
\newblock
\showeprint[arXiv]{2509.02360}
\href{https://doi.org/10.48550/ARXIV.2509.02360}{doi:\nolinkurl{10.48550/ARXIV.2509.02360}}


\bibitem[Guan et~al\mbox{.}(2024)]%
        {verengine}
\bibfield{author}{\bibinfo{person}{Xinyan Guan}, \bibinfo{person}{Yanjiang Liu}, \bibinfo{person}{Xinyu Lu}, \bibinfo{person}{Boxi Cao}, \bibinfo{person}{Ben He}, \bibinfo{person}{Xianpei Han}, \bibinfo{person}{Le Sun}, \bibinfo{person}{Jie Lou}, \bibinfo{person}{Bowen Yu}, \bibinfo{person}{Yaojie Lu}, {and} \bibinfo{person}{Hongyu Lin}.} \bibinfo{year}{2024}\natexlab{}.
\newblock \showarticletitle{Search, Verify and Feedback: Towards Next Generation Post-training Paradigm of Foundation Models via Verifier Engineering}.
\newblock \bibinfo{journal}{\emph{CoRR}}  \bibinfo{volume}{abs/2411.11504} (\bibinfo{year}{2024}).
\newblock
\showeprint[arXiv]{2411.11504}
\href{https://doi.org/10.48550/ARXIV.2411.11504}{doi:\nolinkurl{10.48550/ARXIV.2411.11504}}


\bibitem[He et~al\mbox{.}(2025)]%
        {math_survey}
\bibfield{author}{\bibinfo{person}{Feijuan He}, \bibinfo{person}{Han Lai}, \bibinfo{person}{Jiaqi Liu}, \bibinfo{person}{Bo Wang}, \bibinfo{person}{Haoran Chen}, \bibinfo{person}{Haohan Liu}, {and} \bibinfo{person}{Chenxi Zhang}.} \bibinfo{year}{2025}\natexlab{}.
\newblock \showarticletitle{Solving Mathematical Problems using Large Language Models: {A} Survey}.
\newblock \bibinfo{journal}{\emph{Data Intell.}} \bibinfo{volume}{7}, \bibinfo{number}{4} (\bibinfo{year}{2025}), \bibinfo{pages}{907--946}.
\newblock
\href{https://doi.org/10.3724/2096-7004.DI.2025.0064}{doi:\nolinkurl{10.3724/2096-7004.DI.2025.0064}}


\bibitem[Hong et~al\mbox{.}(2025)]%
        {datainterpreter}
\bibfield{author}{\bibinfo{person}{Sirui Hong}, \bibinfo{person}{Yizhang Lin}, \bibinfo{person}{Bang Liu}, \bibinfo{person}{Bangbang Liu}, \bibinfo{person}{Binhao Wu}, \bibinfo{person}{Ceyao Zhang}, \bibinfo{person}{Danyang Li}, \bibinfo{person}{Jiaqi Chen}, \bibinfo{person}{Jiayi Zhang}, \bibinfo{person}{Jinlin Wang}, \bibinfo{person}{Li Zhang}, \bibinfo{person}{Lingyao Zhang}, \bibinfo{person}{Min Yang}, \bibinfo{person}{Mingchen Zhuge}, \bibinfo{person}{Taicheng Guo}, \bibinfo{person}{Tuo Zhou}, \bibinfo{person}{Wei Tao}, \bibinfo{person}{Robert Tang}, \bibinfo{person}{Xiangtao Lu}, \bibinfo{person}{Xiawu Zheng}, \bibinfo{person}{Xinbing Liang}, \bibinfo{person}{Yaying Fei}, \bibinfo{person}{Yuheng Cheng}, \bibinfo{person}{Yongxin Ni}, \bibinfo{person}{Zhibin Gou}, \bibinfo{person}{Zongze Xu}, \bibinfo{person}{Yuyu Luo}, {and} \bibinfo{person}{Chenglin Wu}.} \bibinfo{year}{2025}\natexlab{}.
\newblock \showarticletitle{Data Interpreter: An {LLM} Agent for Data Science}. In \bibinfo{booktitle}{\emph{Findings of the Association for Computational Linguistics, {ACL} 2025, Vienna, Austria, July 27 - August 1, 2025}}, \bibfield{editor}{\bibinfo{person}{Wanxiang Che}, \bibinfo{person}{Joyce Nabende}, \bibinfo{person}{Ekaterina Shutova}, {and} \bibinfo{person}{Mohammad~Taher Pilehvar}} (Eds.). \bibinfo{publisher}{Association for Computational Linguistics}, \bibinfo{pages}{19796--19821}.
\newblock
\urldef\tempurl%
\url{https://aclanthology.org/2025.findings-acl.1016/}
\showURL{%
\tempurl}


\bibitem[Hu et~al\mbox{.}(2024)]%
        {dabench}
\bibfield{author}{\bibinfo{person}{Xueyu Hu}, \bibinfo{person}{Ziyu Zhao}, \bibinfo{person}{Shuang Wei}, \bibinfo{person}{Ziwei Chai}, \bibinfo{person}{Qianli Ma}, \bibinfo{person}{Guoyin Wang}, \bibinfo{person}{Xuwu Wang}, \bibinfo{person}{Jing Su}, \bibinfo{person}{Jingjing Xu}, \bibinfo{person}{Ming Zhu}, \bibinfo{person}{Yao Cheng}, \bibinfo{person}{Jianbo Yuan}, \bibinfo{person}{Jiwei Li}, \bibinfo{person}{Kun Kuang}, \bibinfo{person}{Yang Yang}, \bibinfo{person}{Hongxia Yang}, {and} \bibinfo{person}{Fei Wu}.} \bibinfo{year}{2024}\natexlab{}.
\newblock \showarticletitle{InfiAgent-DABench: Evaluating Agents on Data Analysis Tasks}. In \bibinfo{booktitle}{\emph{Forty-first International Conference on Machine Learning, {ICML} 2024, Vienna, Austria, July 21-27, 2024}}. \bibinfo{publisher}{OpenReview.net}.
\newblock
\urldef\tempurl%
\url{https://openreview.net/forum?id=d5LURMSfTx}
\showURL{%
\tempurl}


\bibitem[Jing et~al\mbox{.}(2025)]%
        {dsbench}
\bibfield{author}{\bibinfo{person}{Liqiang Jing}, \bibinfo{person}{Zhehui Huang}, \bibinfo{person}{Xiaoyang Wang}, \bibinfo{person}{Wenlin Yao}, \bibinfo{person}{Wenhao Yu}, \bibinfo{person}{Kaixin Ma}, \bibinfo{person}{Hongming Zhang}, \bibinfo{person}{Xinya Du}, {and} \bibinfo{person}{Dong Yu}.} \bibinfo{year}{2025}\natexlab{}.
\newblock \showarticletitle{DSBench: How Far Are Data Science Agents from Becoming Data Science Experts?}. In \bibinfo{booktitle}{\emph{The Thirteenth International Conference on Learning Representations, {ICLR} 2025, Singapore, April 24-28, 2025}}. \bibinfo{publisher}{OpenReview.net}.
\newblock
\urldef\tempurl%
\url{https://openreview.net/forum?id=DSsSPr0RZJ}
\showURL{%
\tempurl}


\bibitem[Khalifa et~al\mbox{.}(2025)]%
        {thinkprm}
\bibfield{author}{\bibinfo{person}{Muhammad Khalifa}, \bibinfo{person}{Rishabh Agarwal}, \bibinfo{person}{Lajanugen Logeswaran}, \bibinfo{person}{Jaekyeom Kim}, \bibinfo{person}{Hao Peng}, \bibinfo{person}{Moontae Lee}, \bibinfo{person}{Honglak Lee}, {and} \bibinfo{person}{Lu Wang}.} \bibinfo{year}{2025}\natexlab{}.
\newblock \showarticletitle{Process Reward Models That Think}.
\newblock \bibinfo{journal}{\emph{CoRR}}  \bibinfo{volume}{abs/2504.16828} (\bibinfo{year}{2025}).
\newblock
\showeprint[arXiv]{2504.16828}
\href{https://doi.org/10.48550/ARXIV.2504.16828}{doi:\nolinkurl{10.48550/ARXIV.2504.16828}}


\bibitem[Li et~al\mbox{.}(2026)]%
        {toolprmbench}
\bibfield{author}{\bibinfo{person}{Dawei Li}, \bibinfo{person}{Yuguang Yao}, \bibinfo{person}{Zhen Tan}, \bibinfo{person}{Huan Liu}, {and} \bibinfo{person}{Ruocheng Guo}.} \bibinfo{year}{2026}\natexlab{}.
\newblock \showarticletitle{ToolPRMBench: Evaluating and Advancing Process Reward Models for Tool-using Agents}.
\newblock \bibinfo{journal}{\emph{arXiv preprint arXiv:2601.12294}} (\bibinfo{year}{2026}).
\newblock


\bibitem[Li et~al\mbox{.}(2025a)]%
        {codeprm}
\bibfield{author}{\bibinfo{person}{Qingyao Li}, \bibinfo{person}{Xinyi Dai}, \bibinfo{person}{Xiangyang Li}, \bibinfo{person}{Weinan Zhang}, \bibinfo{person}{Yasheng Wang}, \bibinfo{person}{Ruiming Tang}, {and} \bibinfo{person}{Yong Yu}.} \bibinfo{year}{2025}\natexlab{a}.
\newblock \showarticletitle{CodePRM: Execution Feedback-enhanced Process Reward Model for Code Generation}. In \bibinfo{booktitle}{\emph{Findings of the Association for Computational Linguistics, {ACL} 2025, Vienna, Austria, July 27 - August 1, 2025}}, \bibfield{editor}{\bibinfo{person}{Wanxiang Che}, \bibinfo{person}{Joyce Nabende}, \bibinfo{person}{Ekaterina Shutova}, {and} \bibinfo{person}{Mohammad~Taher Pilehvar}} (Eds.). \bibinfo{publisher}{Association for Computational Linguistics}, \bibinfo{pages}{8169--8182}.
\newblock
\urldef\tempurl%
\url{https://aclanthology.org/2025.findings-acl.428/}
\showURL{%
\tempurl}


\bibitem[Li et~al\mbox{.}(2025b)]%
        {autosdt}
\bibfield{author}{\bibinfo{person}{Yifei Li}, \bibinfo{person}{Hanane~Nour Moussa}, \bibinfo{person}{Ziru Chen}, \bibinfo{person}{Shijie Chen}, \bibinfo{person}{Botao Yu}, \bibinfo{person}{Mingyi Xue}, \bibinfo{person}{Benjamin Burns}, \bibinfo{person}{Tzu{-}Yao Chiu}, \bibinfo{person}{Vishal Dey}, \bibinfo{person}{Zitong Lu}, \bibinfo{person}{Chen Wei}, \bibinfo{person}{Qianheng Zhang}, \bibinfo{person}{Tianyu Zhang}, \bibinfo{person}{Song Gao}, \bibinfo{person}{Xuhui Huang}, \bibinfo{person}{Xia Ning}, \bibinfo{person}{Nesreen~K. Ahmed}, \bibinfo{person}{Ali Payani}, {and} \bibinfo{person}{Huan Sun}.} \bibinfo{year}{2025}\natexlab{b}.
\newblock \showarticletitle{AutoSDT: Scaling Data-Driven Discovery Tasks Toward Open Co-Scientists}.
\newblock \bibinfo{journal}{\emph{CoRR}}  \bibinfo{volume}{abs/2506.08140} (\bibinfo{year}{2025}).
\newblock
\showeprint[arXiv]{2506.08140}
\href{https://doi.org/10.48550/ARXIV.2506.08140}{doi:\nolinkurl{10.48550/ARXIV.2506.08140}}


\bibitem[Liang et~al\mbox{.}(2026)]%
        {skillnet}
\bibfield{author}{\bibinfo{person}{Yuan Liang}, \bibinfo{person}{Ruobin Zhong}, \bibinfo{person}{Haoming Xu}, \bibinfo{person}{Chen Jiang}, \bibinfo{person}{Yi Zhong}, \bibinfo{person}{Runnan Fang}, \bibinfo{person}{Jia{-}Chen Gu}, \bibinfo{person}{Shumin Deng}, \bibinfo{person}{Yunzhi Yao}, \bibinfo{person}{Mengru Wang}, \bibinfo{person}{Shuofei Qiao}, \bibinfo{person}{Xin Xu}, \bibinfo{person}{Tongtong Wu}, \bibinfo{person}{Kun Wang}, \bibinfo{person}{Yang Liu}, \bibinfo{person}{Zhen Bi}, \bibinfo{person}{Jungang Lou}, \bibinfo{person}{Yuchen~Eleanor Jiang}, \bibinfo{person}{Hangcheng Zhu}, \bibinfo{person}{Gang Yu}, \bibinfo{person}{Haiwen Hong}, \bibinfo{person}{Longtao Huang}, \bibinfo{person}{Hui Xue}, \bibinfo{person}{Chenxi Wang}, \bibinfo{person}{Yijun Wang}, \bibinfo{person}{Zifei Shan}, \bibinfo{person}{Xi Chen}, \bibinfo{person}{Zhaopeng Tu}, \bibinfo{person}{Feiyu Xiong}, \bibinfo{person}{Xin Xie}, \bibinfo{person}{Peng Zhang}, \bibinfo{person}{Zhengke Gui}, \bibinfo{person}{Lei Liang},
  \bibinfo{person}{Jun Zhou}, \bibinfo{person}{Chiyu Wu}, \bibinfo{person}{Jin Shang}, \bibinfo{person}{Yu Gong}, \bibinfo{person}{Junyu Lin}, \bibinfo{person}{Changliang Xu}, \bibinfo{person}{Hongjie Deng}, \bibinfo{person}{Wen Zhang}, \bibinfo{person}{Keyan Ding}, \bibinfo{person}{Qiang Zhang}, \bibinfo{person}{Fei Huang}, \bibinfo{person}{Ningyu Zhang}, \bibinfo{person}{Jeff~Z. Pan}, \bibinfo{person}{Guilin Qi}, \bibinfo{person}{Haofen Wang}, {and} \bibinfo{person}{Huajun Chen}.} \bibinfo{year}{2026}\natexlab{}.
\newblock \showarticletitle{SkillNet: Create, Evaluate, and Connect {AI} Skills}.
\newblock \bibinfo{journal}{\emph{CoRR}}  \bibinfo{volume}{abs/2603.04448} (\bibinfo{year}{2026}).
\newblock
\showeprint[arXiv]{2603.04448}
\href{https://doi.org/10.48550/ARXIV.2603.04448}{doi:\nolinkurl{10.48550/ARXIV.2603.04448}}


\bibitem[Lightman et~al\mbox{.}(2024)]%
        {PRM800K}
\bibfield{author}{\bibinfo{person}{Hunter Lightman}, \bibinfo{person}{Vineet Kosaraju}, \bibinfo{person}{Yuri Burda}, \bibinfo{person}{Harrison Edwards}, \bibinfo{person}{Bowen Baker}, \bibinfo{person}{Teddy Lee}, \bibinfo{person}{Jan Leike}, \bibinfo{person}{John Schulman}, \bibinfo{person}{Ilya Sutskever}, {and} \bibinfo{person}{Karl Cobbe}.} \bibinfo{year}{2024}\natexlab{}.
\newblock \showarticletitle{Let's Verify Step by Step}. In \bibinfo{booktitle}{\emph{The Twelfth International Conference on Learning Representations, {ICLR} 2024, Vienna, Austria, May 7-11, 2024}}. \bibinfo{publisher}{OpenReview.net}.
\newblock
\urldef\tempurl%
\url{https://openreview.net/forum?id=v8L0pN6EOi}
\showURL{%
\tempurl}


\bibitem[Lin et~al\mbox{.}(2025a)]%
        {toolprm}
\bibfield{author}{\bibinfo{person}{Jianghao Lin}, \bibinfo{person}{Yuanyuan Shi}, \bibinfo{person}{Xin Peng}, \bibinfo{person}{Renjie Ding}, \bibinfo{person}{Hairui Wang}, \bibinfo{person}{Yuxuan Peng}, \bibinfo{person}{Bizhe Bai}, \bibinfo{person}{Weixi Song}, \bibinfo{person}{Fengshuo Bai}, \bibinfo{person}{Huacan Chai}, \bibinfo{person}{Weinan Zhang}, \bibinfo{person}{Fei Huang}, {and} \bibinfo{person}{Ying Wen}.} \bibinfo{year}{2025}\natexlab{a}.
\newblock \showarticletitle{ToolPRM: Fine-Grained Inference Scaling of Structured Outputs for Function Calling}.
\newblock \bibinfo{journal}{\emph{CoRR}}  \bibinfo{volume}{abs/2510.14703} (\bibinfo{year}{2025}).
\newblock
\showeprint[arXiv]{2510.14703}
\href{https://doi.org/10.48550/ARXIV.2510.14703}{doi:\nolinkurl{10.48550/ARXIV.2510.14703}}


\bibitem[Lin et~al\mbox{.}(2025b)]%
        {sci_claim_lora}
\bibfield{author}{\bibinfo{person}{Xin Lin}, \bibinfo{person}{Yajiao Wang}, \bibinfo{person}{Zhixiong Zhang}, {and} \bibinfo{person}{Mengting Zhang}.} \bibinfo{year}{2025}\natexlab{b}.
\newblock \showarticletitle{Scientific Claim Recognition via Staged Fine-Tuning with LoRA}.
\newblock \bibinfo{journal}{\emph{Data Intell.}} \bibinfo{volume}{7}, \bibinfo{number}{2} (\bibinfo{year}{2025}), \bibinfo{pages}{303--335}.
\newblock
\href{https://doi.org/10.3724/2096-7004.DI.2025.0009}{doi:\nolinkurl{10.3724/2096-7004.DI.2025.0009}}


\bibitem[Liu et~al\mbox{.}(2025a)]%
        {1btts}
\bibfield{author}{\bibinfo{person}{Runze Liu}, \bibinfo{person}{Junqi Gao}, \bibinfo{person}{Jian Zhao}, \bibinfo{person}{Kaiyan Zhang}, \bibinfo{person}{Xiu Li}, \bibinfo{person}{Biqing Qi}, \bibinfo{person}{Wanli Ouyang}, {and} \bibinfo{person}{Bowen Zhou}.} \bibinfo{year}{2025}\natexlab{a}.
\newblock \showarticletitle{Can 1B {LLM} Surpass 405B LLM? Rethinking Compute-Optimal Test-Time Scaling}.
\newblock \bibinfo{journal}{\emph{CoRR}}  \bibinfo{volume}{abs/2502.06703} (\bibinfo{year}{2025}).
\newblock
\showeprint[arXiv]{2502.06703}
\href{https://doi.org/10.48550/ARXIV.2502.06703}{doi:\nolinkurl{10.48550/ARXIV.2502.06703}}


\bibitem[Liu et~al\mbox{.}(2026a)]%
        {datastorm}
\bibfield{author}{\bibinfo{person}{Shicheng Liu}, \bibinfo{person}{Yucheng Jiang}, \bibinfo{person}{Sajid Farook}, \bibinfo{person}{Camila~Nicollier Sanchez}, \bibinfo{person}{David Fernando~Castro Pena}, {and} \bibinfo{person}{Monica~S. Lam}.} \bibinfo{year}{2026}\natexlab{a}.
\newblock \showarticletitle{DataSTORM: Deep Research on Large-Scale Databases using Exploratory Data Analysis and Data Storytelling}.
\newblock
\urldef\tempurl%
\url{https://api.semanticscholar.org/CorpusID:287248168}
\showURL{%
\tempurl}


\bibitem[Liu et~al\mbox{.}(2026b)]%
        {ddr}
\bibfield{author}{\bibinfo{person}{Wei Liu}, \bibinfo{person}{Peijie Yu}, \bibinfo{person}{Michele Orini}, \bibinfo{person}{Yali Du}, {and} \bibinfo{person}{Yulan He}.} \bibinfo{year}{2026}\natexlab{b}.
\newblock \showarticletitle{Hunt Instead of Wait: Evaluating Deep Data Research on Large Language Models}.
\newblock \bibinfo{journal}{\emph{arXiv preprint arXiv:2602.02039}} (\bibinfo{year}{2026}).
\newblock


\bibitem[Liu et~al\mbox{.}(2025b)]%
        {prm_agentrl}
\bibfield{author}{\bibinfo{person}{Xiaoqian Liu}, \bibinfo{person}{Ke Wang}, \bibinfo{person}{Yuchuan Wu}, \bibinfo{person}{Fei Huang}, \bibinfo{person}{Yongbin Li}, \bibinfo{person}{Junge Zhang}, {and} \bibinfo{person}{Jianbin Jiao}.} \bibinfo{year}{2025}\natexlab{b}.
\newblock \showarticletitle{Agentic Reinforcement Learning with Implicit Step Rewards}.
\newblock \bibinfo{journal}{\emph{CoRR}}  \bibinfo{volume}{abs/2509.19199} (\bibinfo{year}{2025}).
\newblock
\showeprint[arXiv]{2509.19199}
\href{https://doi.org/10.48550/ARXIV.2509.19199}{doi:\nolinkurl{10.48550/ARXIV.2509.19199}}


\bibitem[Lu et~al\mbox{.}(2024)]%
        {ai-scientist}
\bibfield{author}{\bibinfo{person}{Chris Lu}, \bibinfo{person}{Cong Lu}, \bibinfo{person}{Robert~Tjarko Lange}, \bibinfo{person}{Jakob~N. Foerster}, \bibinfo{person}{Jeff Clune}, {and} \bibinfo{person}{David Ha}.} \bibinfo{year}{2024}\natexlab{}.
\newblock \showarticletitle{The {AI} Scientist: Towards Fully Automated Open-Ended Scientific Discovery}.
\newblock \bibinfo{journal}{\emph{CoRR}}  \bibinfo{volume}{abs/2408.06292} (\bibinfo{year}{2024}).
\newblock
\showeprint[arXiv]{2408.06292}
\href{https://doi.org/10.48550/ARXIV.2408.06292}{doi:\nolinkurl{10.48550/ARXIV.2408.06292}}


\bibitem[Luo et~al\mbox{.}(2024)]%
        {omegaprm}
\bibfield{author}{\bibinfo{person}{Liangchen Luo}, \bibinfo{person}{Yinxiao Liu}, \bibinfo{person}{Rosanne Liu}, \bibinfo{person}{Samrat Phatale}, \bibinfo{person}{Harsh Lara}, \bibinfo{person}{Yunxuan Li}, \bibinfo{person}{Lei Shu}, \bibinfo{person}{Yun Zhu}, \bibinfo{person}{Lei Meng}, \bibinfo{person}{Jiao Sun}, {and} \bibinfo{person}{Abhinav Rastogi}.} \bibinfo{year}{2024}\natexlab{}.
\newblock \showarticletitle{Improve Mathematical Reasoning in Language Models by Automated Process Supervision}.
\newblock \bibinfo{journal}{\emph{CoRR}}  \bibinfo{volume}{abs/2406.06592} (\bibinfo{year}{2024}).
\newblock
\showeprint[arXiv]{2406.06592}
\href{https://doi.org/10.48550/ARXIV.2406.06592}{doi:\nolinkurl{10.48550/ARXIV.2406.06592}}


\bibitem[Luo et~al\mbox{.}(2025)]%
        {sft_survey}
\bibfield{author}{\bibinfo{person}{Yitian Luo}, \bibinfo{person}{Yu Liu}, \bibinfo{person}{Lu Zhang}, \bibinfo{person}{Feng Gao}, {and} \bibinfo{person}{Jinguang Gu}.} \bibinfo{year}{2025}\natexlab{}.
\newblock \showarticletitle{A Survey on Quality Evaluation of Instruction Fine-tuning Datasets for Large Language Models}.
\newblock \bibinfo{journal}{\emph{Data Intell.}} \bibinfo{volume}{7}, \bibinfo{number}{3} (\bibinfo{year}{2025}), \bibinfo{pages}{527--566}.
\newblock
\href{https://doi.org/10.3724/2096-7004.DI.2025.0021}{doi:\nolinkurl{10.3724/2096-7004.DI.2025.0021}}


\bibitem[Luong et~al\mbox{.}(2025)]%
        {deepmindmath}
\bibfield{author}{\bibinfo{person}{Thang Luong}, \bibinfo{person}{Dawsen Hwang}, \bibinfo{person}{Hoang~H. Nguyen}, \bibinfo{person}{Golnaz Ghiasi}, \bibinfo{person}{Yuri Chervonyi}, \bibinfo{person}{Insuk Seo}, \bibinfo{person}{Junsu Kim}, \bibinfo{person}{Garrett Bingham}, \bibinfo{person}{Jonathan Lee}, \bibinfo{person}{Swaroop Mishra}, \bibinfo{person}{Alex Zhai}, \bibinfo{person}{Clara~Huiyi Hu}, \bibinfo{person}{Henryk Michalewski}, \bibinfo{person}{Jimin Kim}, \bibinfo{person}{Jeonghyun Ahn}, \bibinfo{person}{Junhwi Bae}, \bibinfo{person}{Xingyou Song}, \bibinfo{person}{Trieu~H. Trinh}, \bibinfo{person}{Quoc~V. Le}, {and} \bibinfo{person}{Junehyuk Jung}.} \bibinfo{year}{2025}\natexlab{}.
\newblock \showarticletitle{Towards Robust Mathematical Reasoning}. In \bibinfo{booktitle}{\emph{Proceedings of the 2025 Conference on Empirical Methods in Natural Language Processing, {EMNLP} 2025, Suzhou, China, November 4-9, 2025}}, \bibfield{editor}{\bibinfo{person}{Christos Christodoulopoulos}, \bibinfo{person}{Tanmoy Chakraborty}, \bibinfo{person}{Carolyn Rose}, {and} \bibinfo{person}{Violet Peng}} (Eds.). \bibinfo{publisher}{Association for Computational Linguistics}, \bibinfo{pages}{35418--35442}.
\newblock
\href{https://doi.org/10.18653/V1/2025.EMNLP-MAIN.1794}{doi:\nolinkurl{10.18653/V1/2025.EMNLP-MAIN.1794}}


\bibitem[Ma et~al\mbox{.}(2023)]%
        {insightpilot}
\bibfield{author}{\bibinfo{person}{Pingchuan Ma}, \bibinfo{person}{Rui Ding}, \bibinfo{person}{Shuai Wang}, \bibinfo{person}{Shi Han}, {and} \bibinfo{person}{Dongmei Zhang}.} \bibinfo{year}{2023}\natexlab{}.
\newblock \showarticletitle{InsightPilot: An LLM-Empowered Automated Data Exploration System}. In \bibinfo{booktitle}{\emph{Proceedings of the 2023 Conference on Empirical Methods in Natural Language Processing, {EMNLP} 2023 - System Demonstrations, Singapore, December 6-10, 2023}}, \bibfield{editor}{\bibinfo{person}{Yansong Feng} {and} \bibinfo{person}{Els Lefever}} (Eds.). \bibinfo{publisher}{Association for Computational Linguistics}, \bibinfo{pages}{346--352}.
\newblock
\href{https://doi.org/10.18653/V1/2023.EMNLP-DEMO.31}{doi:\nolinkurl{10.18653/V1/2023.EMNLP-DEMO.31}}


\bibitem[Nam et~al\mbox{.}(2025)]%
        {dsstar}
\bibfield{author}{\bibinfo{person}{Jaehyun Nam}, \bibinfo{person}{Jinsung Yoon}, \bibinfo{person}{Jiefeng Chen}, {and} \bibinfo{person}{Tomas Pfister}.} \bibinfo{year}{2025}\natexlab{}.
\newblock \showarticletitle{{DS-STAR:} Data Science Agent via Iterative Planning and Verification}.
\newblock \bibinfo{journal}{\emph{CoRR}}  \bibinfo{volume}{abs/2509.21825} (\bibinfo{year}{2025}).
\newblock
\showeprint[arXiv]{2509.21825}
\href{https://doi.org/10.48550/ARXIV.2509.21825}{doi:\nolinkurl{10.48550/ARXIV.2509.21825}}


\bibitem[Ni et~al\mbox{.}(2026)]%
        {trace2skill}
\bibfield{author}{\bibinfo{person}{Jingwei Ni}, \bibinfo{person}{Yihao Liu}, \bibinfo{person}{Xinpeng Liu}, \bibinfo{person}{Yutao Sun}, \bibinfo{person}{Mengyu Zhou}, \bibinfo{person}{Pengyu Cheng}, \bibinfo{person}{Dexin Wang}, \bibinfo{person}{Erchao Zhao}, \bibinfo{person}{Xiaoxi Jiang}, {and} \bibinfo{person}{Guanjun Jiang}.} \bibinfo{year}{2026}\natexlab{}.
\newblock \showarticletitle{Trace2Skill: Distill Trajectory-Local Lessons into Transferable Agent Skills}.
\newblock \bibinfo{journal}{\emph{CoRR}}  \bibinfo{volume}{abs/2603.25158} (\bibinfo{year}{2026}).
\newblock
\showeprint[arXiv]{2603.25158}
\href{https://doi.org/10.48550/ARXIV.2603.25158}{doi:\nolinkurl{10.48550/ARXIV.2603.25158}}


\bibitem[Nie et~al\mbox{.}(2026)]%
        {dsgym}
\bibfield{author}{\bibinfo{person}{Fan Nie}, \bibinfo{person}{Junlin Wang}, \bibinfo{person}{Harper Hua}, \bibinfo{person}{Federico Bianchi}, \bibinfo{person}{Yongchan Kwon}, \bibinfo{person}{Zhenting Qi}, \bibinfo{person}{Owen Queen}, \bibinfo{person}{Shang Zhu}, {and} \bibinfo{person}{James Zou}.} \bibinfo{year}{2026}\natexlab{}.
\newblock \showarticletitle{DSGym: A Holistic Framework for Evaluating and Training Data Science Agents}.
\newblock \bibinfo{journal}{\emph{arXiv preprint arXiv:2601.16344}} (\bibinfo{year}{2026}).
\newblock


\bibitem[Qi et~al\mbox{.}(2026)]%
        {datacross}
\bibfield{author}{\bibinfo{person}{Ruyi Qi}, \bibinfo{person}{Zhou Liu}, {and} \bibinfo{person}{Wentao Zhang}.} \bibinfo{year}{2026}\natexlab{}.
\newblock \showarticletitle{DataCross: A Unified Benchmark and Agent Framework for Cross-Modal Heterogeneous Data Analysis}.
\newblock \bibinfo{journal}{\emph{ArXiv}}  \bibinfo{volume}{abs/2601.21403} (\bibinfo{year}{2026}).
\newblock
\urldef\tempurl%
\url{https://api.semanticscholar.org/CorpusID:285140426}
\showURL{%
\tempurl}


\bibitem[Qian et~al\mbox{.}(2025)]%
        {toolrl}
\bibfield{author}{\bibinfo{person}{Cheng Qian}, \bibinfo{person}{Emre~Can Acikgoz}, \bibinfo{person}{Qi He}, \bibinfo{person}{Hongru Wang}, \bibinfo{person}{Xiusi Chen}, \bibinfo{person}{Dilek Hakkani{-}T{\"{u}}r}, \bibinfo{person}{Gokhan Tur}, {and} \bibinfo{person}{Heng Ji}.} \bibinfo{year}{2025}\natexlab{}.
\newblock \showarticletitle{ToolRL: Reward is All Tool Learning Needs}.
\newblock \bibinfo{journal}{\emph{CoRR}}  \bibinfo{volume}{abs/2504.13958} (\bibinfo{year}{2025}).
\newblock
\showeprint[arXiv]{2504.13958}
\href{https://doi.org/10.48550/ARXIV.2504.13958}{doi:\nolinkurl{10.48550/ARXIV.2504.13958}}


\bibitem[Qiao et~al\mbox{.}(2023)]%
        {reasoning-survey}
\bibfield{author}{\bibinfo{person}{Shuofei Qiao}, \bibinfo{person}{Yixin Ou}, \bibinfo{person}{Ningyu Zhang}, \bibinfo{person}{Xiang Chen}, \bibinfo{person}{Yunzhi Yao}, \bibinfo{person}{Shumin Deng}, \bibinfo{person}{Chuanqi Tan}, \bibinfo{person}{Fei Huang}, {and} \bibinfo{person}{Huajun Chen}.} \bibinfo{year}{2023}\natexlab{}.
\newblock \showarticletitle{Reasoning with Language Model Prompting: {A} Survey}. In \bibinfo{booktitle}{\emph{Proceedings of the 61st Annual Meeting of the Association for Computational Linguistics (Volume 1: Long Papers), {ACL} 2023, Toronto, Canada, July 9-14, 2023}}, \bibfield{editor}{\bibinfo{person}{Anna Rogers}, \bibinfo{person}{Jordan~L. Boyd{-}Graber}, {and} \bibinfo{person}{Naoaki Okazaki}} (Eds.). \bibinfo{publisher}{Association for Computational Linguistics}, \bibinfo{pages}{5368--5393}.
\newblock
\href{https://doi.org/10.18653/V1/2023.ACL-LONG.294}{doi:\nolinkurl{10.18653/V1/2023.ACL-LONG.294}}


\bibitem[Qiao et~al\mbox{.}(2025)]%
        {datamind}
\bibfield{author}{\bibinfo{person}{Shuofei Qiao}, \bibinfo{person}{Yanqiu Zhao}, \bibinfo{person}{Zhisong Qiu}, \bibinfo{person}{Xiaobin Wang}, \bibinfo{person}{Jintian Zhang}, \bibinfo{person}{Zhao Bin}, \bibinfo{person}{Ningyu Zhang}, \bibinfo{person}{Yong Jiang}, \bibinfo{person}{Pengjun Xie}, \bibinfo{person}{Fei Huang}, {and} \bibinfo{person}{Huajun Chen}.} \bibinfo{year}{2025}\natexlab{}.
\newblock \showarticletitle{Scaling Generalist Data-Analytic Agents}.
\newblock \bibinfo{journal}{\emph{CoRR}}  \bibinfo{volume}{abs/2509.25084} (\bibinfo{year}{2025}).
\newblock
\showeprint[arXiv]{2509.25084}
\href{https://doi.org/10.48550/ARXIV.2509.25084}{doi:\nolinkurl{10.48550/ARXIV.2509.25084}}


\bibitem[Rahman et~al\mbox{.}(2025)]%
        {spark}
\bibfield{author}{\bibinfo{person}{Salman Rahman}, \bibinfo{person}{Sruthi Gorantla}, \bibinfo{person}{Arpit Gupta}, \bibinfo{person}{Swastik Roy}, \bibinfo{person}{Nanyun Peng}, {and} \bibinfo{person}{Yang Liu}.} \bibinfo{year}{2025}\natexlab{}.
\newblock \showarticletitle{{SPARK:} Stepwise Process-Aware Rewards for Reference-Free Reinforcement Learning}.
\newblock \bibinfo{journal}{\emph{CoRR}}  \bibinfo{volume}{abs/2512.03244} (\bibinfo{year}{2025}).
\newblock
\showeprint[arXiv]{2512.03244}
\href{https://doi.org/10.48550/ARXIV.2512.03244}{doi:\nolinkurl{10.48550/ARXIV.2512.03244}}


\bibitem[Ren et~al\mbox{.}(2025)]%
        {deepseek_prover}
\bibfield{author}{\bibinfo{person}{Z.~Z. Ren}, \bibinfo{person}{Zhihong Shao}, \bibinfo{person}{Junxiao Song}, \bibinfo{person}{Huajian Xin}, \bibinfo{person}{Haocheng Wang}, \bibinfo{person}{Wanjia Zhao}, \bibinfo{person}{Liyue Zhang}, \bibinfo{person}{Zhe Fu}, \bibinfo{person}{Qihao Zhu}, \bibinfo{person}{Dejian Yang}, \bibinfo{person}{Z.~F. Wu}, \bibinfo{person}{Zhibin Gou}, \bibinfo{person}{Shirong Ma}, \bibinfo{person}{Hongxuan Tang}, \bibinfo{person}{Yuxuan Liu}, \bibinfo{person}{Wenjun Gao}, \bibinfo{person}{Daya Guo}, {and} \bibinfo{person}{Chong Ruan}.} \bibinfo{year}{2025}\natexlab{}.
\newblock \showarticletitle{DeepSeek-Prover-V2: Advancing Formal Mathematical Reasoning via Reinforcement Learning for Subgoal Decomposition}.
\newblock \bibinfo{journal}{\emph{CoRR}}  \bibinfo{volume}{abs/2504.21801} (\bibinfo{year}{2025}).
\newblock
\showeprint[arXiv]{2504.21801}
\href{https://doi.org/10.48550/ARXIV.2504.21801}{doi:\nolinkurl{10.48550/ARXIV.2504.21801}}


\bibitem[Schmidgall et~al\mbox{.}(2025)]%
        {agent_laboratory}
\bibfield{author}{\bibinfo{person}{Samuel Schmidgall}, \bibinfo{person}{Yusheng Su}, \bibinfo{person}{Ze Wang}, \bibinfo{person}{Ximeng Sun}, \bibinfo{person}{Jialian Wu}, \bibinfo{person}{Xiaodong Yu}, \bibinfo{person}{Jiang Liu}, \bibinfo{person}{Zicheng Liu}, {and} \bibinfo{person}{Emad Barsoum}.} \bibinfo{year}{2025}\natexlab{}.
\newblock \showarticletitle{Agent Laboratory: Using {LLM} Agents as Research Assistants}.
\newblock \bibinfo{journal}{\emph{CoRR}}  \bibinfo{volume}{abs/2501.04227} (\bibinfo{year}{2025}).
\newblock
\showeprint[arXiv]{2501.04227}
\href{https://doi.org/10.48550/ARXIV.2501.04227}{doi:\nolinkurl{10.48550/ARXIV.2501.04227}}


\bibitem[Setlur et~al\mbox{.}(2025)]%
        {pav}
\bibfield{author}{\bibinfo{person}{Amrith Setlur}, \bibinfo{person}{Chirag Nagpal}, \bibinfo{person}{Adam Fisch}, \bibinfo{person}{Xinyang Geng}, \bibinfo{person}{Jacob Eisenstein}, \bibinfo{person}{Rishabh Agarwal}, \bibinfo{person}{Alekh Agarwal}, \bibinfo{person}{Jonathan Berant}, {and} \bibinfo{person}{Aviral Kumar}.} \bibinfo{year}{2025}\natexlab{}.
\newblock \showarticletitle{Rewarding Progress: Scaling Automated Process Verifiers for {LLM} Reasoning}. In \bibinfo{booktitle}{\emph{The Thirteenth International Conference on Learning Representations, {ICLR} 2025, Singapore, April 24-28, 2025}}. \bibinfo{publisher}{OpenReview.net}.
\newblock
\urldef\tempurl%
\url{https://openreview.net/forum?id=A6Y7AqlzLW}
\showURL{%
\tempurl}


\bibitem[Shao et~al\mbox{.}(2025)]%
        {deepseekmathv2}
\bibfield{author}{\bibinfo{person}{Zhihong Shao}, \bibinfo{person}{Yuxiang Luo}, \bibinfo{person}{Chengda Lu}, \bibinfo{person}{Z.~Z. Ren}, \bibinfo{person}{Jiewen Hu}, \bibinfo{person}{Tian Ye}, \bibinfo{person}{Zhibin Gou}, \bibinfo{person}{Shirong Ma}, {and} \bibinfo{person}{Xiaokang Zhang}.} \bibinfo{year}{2025}\natexlab{}.
\newblock \showarticletitle{DeepSeekMath-V2: Towards Self-Verifiable Mathematical Reasoning}.
\newblock \bibinfo{journal}{\emph{CoRR}}  \bibinfo{volume}{abs/2511.22570} (\bibinfo{year}{2025}).
\newblock
\showeprint[arXiv]{2511.22570}
\href{https://doi.org/10.48550/ARXIV.2511.22570}{doi:\nolinkurl{10.48550/ARXIV.2511.22570}}


\bibitem[Shao et~al\mbox{.}(2024)]%
        {grpo}
\bibfield{author}{\bibinfo{person}{Zhihong Shao}, \bibinfo{person}{Peiyi Wang}, \bibinfo{person}{Qihao Zhu}, \bibinfo{person}{Runxin Xu}, \bibinfo{person}{Junxiao Song}, \bibinfo{person}{Mingchuan Zhang}, \bibinfo{person}{Y.~K. Li}, \bibinfo{person}{Y. Wu}, {and} \bibinfo{person}{Daya Guo}.} \bibinfo{year}{2024}\natexlab{}.
\newblock \showarticletitle{DeepSeekMath: Pushing the Limits of Mathematical Reasoning in Open Language Models}.
\newblock \bibinfo{journal}{\emph{CoRR}}  \bibinfo{volume}{abs/2402.03300} (\bibinfo{year}{2024}).
\newblock
\showeprint[arXiv]{2402.03300}
\href{https://doi.org/10.48550/ARXIV.2402.03300}{doi:\nolinkurl{10.48550/ARXIV.2402.03300}}


\bibitem[Sheng et~al\mbox{.}(2025)]%
        {verl}
\bibfield{author}{\bibinfo{person}{Guangming Sheng}, \bibinfo{person}{Chi Zhang}, \bibinfo{person}{Zilingfeng Ye}, \bibinfo{person}{Xibin Wu}, \bibinfo{person}{Wang Zhang}, \bibinfo{person}{Ru Zhang}, \bibinfo{person}{Yanghua Peng}, \bibinfo{person}{Haibin Lin}, {and} \bibinfo{person}{Chuan Wu}.} \bibinfo{year}{2025}\natexlab{}.
\newblock \showarticletitle{HybridFlow: {A} Flexible and Efficient {RLHF} Framework}. In \bibinfo{booktitle}{\emph{Proceedings of the Twentieth European Conference on Computer Systems, EuroSys 2025, Rotterdam, The Netherlands, 30 March 2025 - 3 April 2025}}. \bibinfo{publisher}{{ACM}}, \bibinfo{pages}{1279--1297}.
\newblock
\href{https://doi.org/10.1145/3689031.3696075}{doi:\nolinkurl{10.1145/3689031.3696075}}


\bibitem[Snell et~al\mbox{.}(2024)]%
        {tts}
\bibfield{author}{\bibinfo{person}{Charlie Snell}, \bibinfo{person}{Jaehoon Lee}, \bibinfo{person}{Kelvin Xu}, {and} \bibinfo{person}{Aviral Kumar}.} \bibinfo{year}{2024}\natexlab{}.
\newblock \showarticletitle{Scaling {LLM} Test-Time Compute Optimally can be More Effective than Scaling Model Parameters}.
\newblock \bibinfo{journal}{\emph{CoRR}}  \bibinfo{volume}{abs/2408.03314} (\bibinfo{year}{2024}).
\newblock
\showeprint[arXiv]{2408.03314}
\href{https://doi.org/10.48550/ARXIV.2408.03314}{doi:\nolinkurl{10.48550/ARXIV.2408.03314}}


\bibitem[Sun et~al\mbox{.}(2025)]%
        {agenticdata}
\bibfield{author}{\bibinfo{person}{Ji Sun}, \bibinfo{person}{Guoliang Li}, \bibinfo{person}{Peiyao Zhou}, \bibinfo{person}{Yihui Ma}, \bibinfo{person}{Jingzhe Xu}, {and} \bibinfo{person}{Yuan Li}.} \bibinfo{year}{2025}\natexlab{}.
\newblock \showarticletitle{AgenticData: An Agentic Data Analytics System for Heterogeneous Data}.
\newblock \bibinfo{journal}{\emph{CoRR}}  \bibinfo{volume}{abs/2508.05002} (\bibinfo{year}{2025}).
\newblock
\showeprint[arXiv]{2508.05002}
\href{https://doi.org/10.48550/ARXIV.2508.05002}{doi:\nolinkurl{10.48550/ARXIV.2508.05002}}


\bibitem[Tang et~al\mbox{.}(2025)]%
        {exploretableprm}
\bibfield{author}{\bibinfo{person}{Lei Tang}, \bibinfo{person}{Wei Zhou}, {and} \bibinfo{person}{Mohsen Mesgar}.} \bibinfo{year}{2025}\natexlab{}.
\newblock \showarticletitle{Exploring Generative Process Reward Modeling for Semi-Structured Data: {A} Case Study of Table Question Answering}.
\newblock \bibinfo{journal}{\emph{CoRR}}  \bibinfo{volume}{abs/2510.20304} (\bibinfo{year}{2025}).
\newblock
\showeprint[arXiv]{2510.20304}
\href{https://doi.org/10.48550/ARXIV.2510.20304}{doi:\nolinkurl{10.48550/ARXIV.2510.20304}}


\bibitem[Wang et~al\mbox{.}(2026)]%
        {skillx}
\bibfield{author}{\bibinfo{person}{Chenxi Wang}, \bibinfo{person}{Zhuoyun Yu}, \bibinfo{person}{Xinghong Xie}, \bibinfo{person}{Wuguannan Yao}, \bibinfo{person}{Runnan Fang}, \bibinfo{person}{Shuofei Qiao}, \bibinfo{person}{Kexin Cao}, \bibinfo{person}{Guozhou Zheng}, \bibinfo{person}{Xiang Qi}, \bibinfo{person}{Peng Zhang}, {and} \bibinfo{person}{Shumin Deng}.} \bibinfo{year}{2026}\natexlab{}.
\newblock \showarticletitle{SkillX: Automatically Constructing Skill Knowledge Bases for Agents}.
\newblock
\urldef\tempurl%
\url{https://api.semanticscholar.org/CorpusID:287204111}
\showURL{%
\tempurl}


\bibitem[Wang et~al\mbox{.}(2024)]%
        {math-shepherd}
\bibfield{author}{\bibinfo{person}{Peiyi Wang}, \bibinfo{person}{Lei Li}, \bibinfo{person}{Zhihong Shao}, \bibinfo{person}{Runxin Xu}, \bibinfo{person}{Damai Dai}, \bibinfo{person}{Yifei Li}, \bibinfo{person}{Deli Chen}, \bibinfo{person}{Yu Wu}, {and} \bibinfo{person}{Zhifang Sui}.} \bibinfo{year}{2024}\natexlab{}.
\newblock \showarticletitle{Math-Shepherd: Verify and Reinforce LLMs Step-by-step without Human Annotations}. In \bibinfo{booktitle}{\emph{Proceedings of the 62nd Annual Meeting of the Association for Computational Linguistics (Volume 1: Long Papers), {ACL} 2024, Bangkok, Thailand, August 11-16, 2024}}, \bibfield{editor}{\bibinfo{person}{Lun{-}Wei Ku}, \bibinfo{person}{Andre Martins}, {and} \bibinfo{person}{Vivek Srikumar}} (Eds.). \bibinfo{publisher}{Association for Computational Linguistics}, \bibinfo{pages}{9426--9439}.
\newblock
\href{https://doi.org/10.18653/V1/2024.ACL-LONG.510}{doi:\nolinkurl{10.18653/V1/2024.ACL-LONG.510}}


\bibitem[Wang et~al\mbox{.}(2025)]%
        {dsagent_survey}
\bibfield{author}{\bibinfo{person}{Peiran Wang}, \bibinfo{person}{Yaoning Yu}, \bibinfo{person}{Ke Chen}, \bibinfo{person}{Xianyang Zhan}, {and} \bibinfo{person}{Haohan Wang}.} \bibinfo{year}{2025}\natexlab{}.
\newblock \showarticletitle{Large Language Model-based Data Science Agent: {A} Survey}.
\newblock \bibinfo{journal}{\emph{CoRR}}  \bibinfo{volume}{abs/2508.02744} (\bibinfo{year}{2025}).
\newblock
\showeprint[arXiv]{2508.02744}
\href{https://doi.org/10.48550/ARXIV.2508.02744}{doi:\nolinkurl{10.48550/ARXIV.2508.02744}}


\bibitem[Wen et~al\mbox{.}(2026)]%
        {smartsearch}
\bibfield{author}{\bibinfo{person}{Tongyu Wen}, \bibinfo{person}{Guanting Dong}, {and} \bibinfo{person}{Zhicheng Dou}.} \bibinfo{year}{2026}\natexlab{}.
\newblock \showarticletitle{SmartSearch: Process Reward-Guided Query Refinement for Search Agents}.
\newblock \bibinfo{journal}{\emph{arXiv preprint arXiv:2601.04888}} (\bibinfo{year}{2026}).
\newblock


\bibitem[Wu et~al\mbox{.}(2025)]%
        {tablebench}
\bibfield{author}{\bibinfo{person}{Xianjie Wu}, \bibinfo{person}{Jian Yang}, \bibinfo{person}{Linzheng Chai}, \bibinfo{person}{Ge Zhang}, \bibinfo{person}{Jiaheng Liu}, \bibinfo{person}{Xeron Du}, \bibinfo{person}{Di Liang}, \bibinfo{person}{Daixin Shu}, \bibinfo{person}{Xianfu Cheng}, \bibinfo{person}{Tianzhen Sun}, \bibinfo{person}{Tongliang Li}, \bibinfo{person}{Zhoujun Li}, {and} \bibinfo{person}{Guanglin Niu}.} \bibinfo{year}{2025}\natexlab{}.
\newblock \showarticletitle{TableBench: {A} Comprehensive and Complex Benchmark for Table Question Answering}. In \bibinfo{booktitle}{\emph{AAAI-25, Sponsored by the Association for the Advancement of Artificial Intelligence, February 25 - March 4, 2025, Philadelphia, PA, {USA}}}, \bibfield{editor}{\bibinfo{person}{Toby Walsh}, \bibinfo{person}{Julie Shah}, {and} \bibinfo{person}{Zico Kolter}} (Eds.). \bibinfo{publisher}{{AAAI} Press}, \bibinfo{pages}{25497--25506}.
\newblock
\href{https://doi.org/10.1609/AAAI.V39I24.34739}{doi:\nolinkurl{10.1609/AAAI.V39I24.34739}}


\bibitem[Xi et~al\mbox{.}(2025)]%
        {agentprm}
\bibfield{author}{\bibinfo{person}{Zhiheng Xi}, \bibinfo{person}{Chenyang Liao}, \bibinfo{person}{Guanyu Li}, \bibinfo{person}{Yajie Yang}, \bibinfo{person}{Wenxiang Chen}, \bibinfo{person}{Zhihao Zhang}, \bibinfo{person}{Binghai Wang}, \bibinfo{person}{Senjie Jin}, \bibinfo{person}{Yuhao Zhou}, \bibinfo{person}{Jian Guan}, \bibinfo{person}{Wei Wu}, \bibinfo{person}{Tao Ji}, \bibinfo{person}{Tao Gui}, \bibinfo{person}{Qi Zhang}, {and} \bibinfo{person}{Xuanjing Huang}.} \bibinfo{year}{2025}\natexlab{}.
\newblock \showarticletitle{AgentPRM: Process Reward Models for {LLM} Agents via Step-Wise Promise and Progress}.
\newblock \bibinfo{journal}{\emph{CoRR}}  \bibinfo{volume}{abs/2511.08325} (\bibinfo{year}{2025}).
\newblock
\showeprint[arXiv]{2511.08325}
\href{https://doi.org/10.48550/ARXIV.2511.08325}{doi:\nolinkurl{10.48550/ARXIV.2511.08325}}


\bibitem[Xu et~al\mbox{.}(2025)]%
        {dagent}
\bibfield{author}{\bibinfo{person}{Wenyi Xu}, \bibinfo{person}{Yuren Mao}, \bibinfo{person}{Xiaolu Zhang}, \bibinfo{person}{Chao Zhang}, \bibinfo{person}{Xuemei Dong}, \bibinfo{person}{Mengfei Zhang}, {and} \bibinfo{person}{Yunjun Gao}.} \bibinfo{year}{2025}\natexlab{}.
\newblock \showarticletitle{DAgent: {A} Relational Database-Driven Data Analysis Report Generation Agent}.
\newblock \bibinfo{journal}{\emph{CoRR}}  \bibinfo{volume}{abs/2503.13269} (\bibinfo{year}{2025}).
\newblock
\showeprint[arXiv]{2503.13269}
\href{https://doi.org/10.48550/ARXIV.2503.13269}{doi:\nolinkurl{10.48550/ARXIV.2503.13269}}


\bibitem[Yang et~al\mbox{.}(2025)]%
        {qwen3}
\bibfield{author}{\bibinfo{person}{An Yang}, \bibinfo{person}{Anfeng Li}, \bibinfo{person}{Baosong Yang}, \bibinfo{person}{Beichen Zhang}, \bibinfo{person}{Binyuan Hui}, \bibinfo{person}{Bo Zheng}, \bibinfo{person}{Bowen Yu}, \bibinfo{person}{Chang Gao}, \bibinfo{person}{Chengen Huang}, \bibinfo{person}{Chenxu Lv}, \bibinfo{person}{Chujie Zheng}, \bibinfo{person}{Dayiheng Liu}, \bibinfo{person}{Fan Zhou}, \bibinfo{person}{Fei Huang}, \bibinfo{person}{Feng Hu}, \bibinfo{person}{Hao Ge}, \bibinfo{person}{Haoran Wei}, \bibinfo{person}{Huan Lin}, \bibinfo{person}{Jialong Tang}, \bibinfo{person}{Jian Yang}, \bibinfo{person}{Jianhong Tu}, \bibinfo{person}{Jianwei Zhang}, \bibinfo{person}{Jian Yang}, \bibinfo{person}{Jiaxi Yang}, \bibinfo{person}{Jingren Zhou}, \bibinfo{person}{Junyang Lin}, \bibinfo{person}{Kai Dang}, \bibinfo{person}{Keqin Bao}, \bibinfo{person}{Kexin Yang}, \bibinfo{person}{Le Yu}, \bibinfo{person}{Lianghao Deng}, \bibinfo{person}{Mei Li}, \bibinfo{person}{Mingfeng Xue}, \bibinfo{person}{Mingze
  Li}, \bibinfo{person}{Pei Zhang}, \bibinfo{person}{Peng Wang}, \bibinfo{person}{Qin Zhu}, \bibinfo{person}{Rui Men}, \bibinfo{person}{Ruize Gao}, \bibinfo{person}{Shixuan Liu}, \bibinfo{person}{Shuang Luo}, \bibinfo{person}{Tianhao Li}, \bibinfo{person}{Tianyi Tang}, \bibinfo{person}{Wenbiao Yin}, \bibinfo{person}{Xingzhang Ren}, \bibinfo{person}{Xinyu Wang}, \bibinfo{person}{Xinyu Zhang}, \bibinfo{person}{Xuancheng Ren}, \bibinfo{person}{Yang Fan}, \bibinfo{person}{Yang Su}, \bibinfo{person}{Yichang Zhang}, \bibinfo{person}{Yinger Zhang}, \bibinfo{person}{Yu Wan}, \bibinfo{person}{Yuqiong Liu}, \bibinfo{person}{Zekun Wang}, \bibinfo{person}{Zeyu Cui}, \bibinfo{person}{Zhenru Zhang}, \bibinfo{person}{Zhipeng Zhou}, {and} \bibinfo{person}{Zihan Qiu}.} \bibinfo{year}{2025}\natexlab{}.
\newblock \showarticletitle{Qwen3 Technical Report}.
\newblock \bibinfo{journal}{\emph{CoRR}}  \bibinfo{volume}{abs/2505.09388} (\bibinfo{year}{2025}).
\newblock
\showeprint[arXiv]{2505.09388}
\href{https://doi.org/10.48550/ARXIV.2505.09388}{doi:\nolinkurl{10.48550/ARXIV.2505.09388}}


\bibitem[Yang et~al\mbox{.}(2024)]%
        {matplotagent}
\bibfield{author}{\bibinfo{person}{Zhiyu Yang}, \bibinfo{person}{Zihan Zhou}, \bibinfo{person}{Shuo Wang}, \bibinfo{person}{Xin Cong}, \bibinfo{person}{Xu Han}, \bibinfo{person}{Yukun Yan}, \bibinfo{person}{Zhenghao Liu}, \bibinfo{person}{Zhixing Tan}, \bibinfo{person}{Pengyuan Liu}, \bibinfo{person}{Dong Yu}, \bibinfo{person}{Zhiyuan Liu}, \bibinfo{person}{Xiaodong Shi}, {and} \bibinfo{person}{Maosong Sun}.} \bibinfo{year}{2024}\natexlab{}.
\newblock \showarticletitle{MatPlotAgent: Method and Evaluation for LLM-Based Agentic Scientific Data Visualization}. In \bibinfo{booktitle}{\emph{Findings of the Association for Computational Linguistics, {ACL} 2024, Bangkok, Thailand and virtual meeting, August 11-16, 2024}}, \bibfield{editor}{\bibinfo{person}{Lun{-}Wei Ku}, \bibinfo{person}{Andre Martins}, {and} \bibinfo{person}{Vivek Srikumar}} (Eds.). \bibinfo{publisher}{Association for Computational Linguistics}, \bibinfo{pages}{11789--11804}.
\newblock
\href{https://doi.org/10.18653/V1/2024.FINDINGS-ACL.701}{doi:\nolinkurl{10.18653/V1/2024.FINDINGS-ACL.701}}


\bibitem[Yao et~al\mbox{.}(2023)]%
        {react}
\bibfield{author}{\bibinfo{person}{Shunyu Yao}, \bibinfo{person}{Jeffrey Zhao}, \bibinfo{person}{Dian Yu}, \bibinfo{person}{Nan Du}, \bibinfo{person}{Izhak Shafran}, \bibinfo{person}{Karthik~R. Narasimhan}, {and} \bibinfo{person}{Yuan Cao}.} \bibinfo{year}{2023}\natexlab{}.
\newblock \showarticletitle{ReAct: Synergizing Reasoning and Acting in Language Models}. In \bibinfo{booktitle}{\emph{The Eleventh International Conference on Learning Representations, {ICLR} 2023, Kigali, Rwanda, May 1-5, 2023}}. \bibinfo{publisher}{OpenReview.net}.
\newblock
\urldef\tempurl%
\url{https://openreview.net/forum?id=WE\_vluYUL-X}
\showURL{%
\tempurl}


\bibitem[You et~al\mbox{.}(2025)]%
        {datawiseagent}
\bibfield{author}{\bibinfo{person}{Ziming You}, \bibinfo{person}{Yumiao Zhang}, \bibinfo{person}{Dexuan Xu}, \bibinfo{person}{Yiwei Lou}, \bibinfo{person}{Yandong Yan}, \bibinfo{person}{Wei Wang}, \bibinfo{person}{Huaming Zhang}, {and} \bibinfo{person}{Yu Huang}.} \bibinfo{year}{2025}\natexlab{}.
\newblock \showarticletitle{DatawiseAgent: {A} Notebook-Centric {LLM} Agent Framework for Automated Data Science}.
\newblock \bibinfo{journal}{\emph{CoRR}}  \bibinfo{volume}{abs/2503.07044} (\bibinfo{year}{2025}).
\newblock
\showeprint[arXiv]{2503.07044}
\href{https://doi.org/10.48550/ARXIV.2503.07044}{doi:\nolinkurl{10.48550/ARXIV.2503.07044}}


\bibitem[Yu et~al\mbox{.}(2025)]%
        {dapo}
\bibfield{author}{\bibinfo{person}{Qiying Yu}, \bibinfo{person}{Zheng Zhang}, \bibinfo{person}{Ruofei Zhu}, \bibinfo{person}{Yufeng Yuan}, \bibinfo{person}{Xiaochen Zuo}, \bibinfo{person}{Yu Yue}, \bibinfo{person}{Tiantian Fan}, \bibinfo{person}{Gaohong Liu}, \bibinfo{person}{Lingjun Liu}, \bibinfo{person}{Xin Liu}, \bibinfo{person}{Haibin Lin}, \bibinfo{person}{Zhiqi Lin}, \bibinfo{person}{Bole Ma}, \bibinfo{person}{Guangming Sheng}, \bibinfo{person}{Yuxuan Tong}, \bibinfo{person}{Chi Zhang}, \bibinfo{person}{Mofan Zhang}, \bibinfo{person}{Wang Zhang}, \bibinfo{person}{Hang Zhu}, \bibinfo{person}{Jinhua Zhu}, \bibinfo{person}{Jiaze Chen}, \bibinfo{person}{Jiangjie Chen}, \bibinfo{person}{Chengyi Wang}, \bibinfo{person}{Hongli Yu}, \bibinfo{person}{Weinan Dai}, \bibinfo{person}{Yuxuan Song}, \bibinfo{person}{Xiangpeng Wei}, \bibinfo{person}{Hao Zhou}, \bibinfo{person}{Jingjing Liu}, \bibinfo{person}{Wei{-}Ying Ma}, \bibinfo{person}{Ya{-}Qin Zhang}, \bibinfo{person}{Lin Yan}, \bibinfo{person}{Mu Qiao},
  \bibinfo{person}{Yonghui Wu}, {and} \bibinfo{person}{Mingxuan Wang}.} \bibinfo{year}{2025}\natexlab{}.
\newblock \showarticletitle{{DAPO:} An Open-Source {LLM} Reinforcement Learning System at Scale}.
\newblock \bibinfo{journal}{\emph{CoRR}}  \bibinfo{volume}{abs/2503.14476} (\bibinfo{year}{2025}).
\newblock
\showeprint[arXiv]{2503.14476}
\href{https://doi.org/10.48550/ARXIV.2503.14476}{doi:\nolinkurl{10.48550/ARXIV.2503.14476}}


\bibitem[Yu et~al\mbox{.}(2024)]%
        {orps}
\bibfield{author}{\bibinfo{person}{Zhuohao Yu}, \bibinfo{person}{Weizheng Gu}, \bibinfo{person}{Yidong Wang}, \bibinfo{person}{Zhengran Zeng}, \bibinfo{person}{Jindong Wang}, \bibinfo{person}{Wei Ye}, {and} \bibinfo{person}{Shikun Zhang}.} \bibinfo{year}{2024}\natexlab{}.
\newblock \showarticletitle{Outcome-Refining Process Supervision for Code Generation}.
\newblock \bibinfo{journal}{\emph{CoRR}}  \bibinfo{volume}{abs/2412.15118} (\bibinfo{year}{2024}).
\newblock
\showeprint[arXiv]{2412.15118}
\href{https://doi.org/10.48550/ARXIV.2412.15118}{doi:\nolinkurl{10.48550/ARXIV.2412.15118}}


\bibitem[Yuan et~al\mbox{.}(2024)]%
        {self_rewarding_rm}
\bibfield{author}{\bibinfo{person}{Weizhe Yuan}, \bibinfo{person}{Richard~Yuanzhe Pang}, \bibinfo{person}{Kyunghyun Cho}, \bibinfo{person}{Xian Li}, \bibinfo{person}{Sainbayar Sukhbaatar}, \bibinfo{person}{Jing Xu}, {and} \bibinfo{person}{Jason Weston}.} \bibinfo{year}{2024}\natexlab{}.
\newblock \showarticletitle{Self-Rewarding Language Models}. In \bibinfo{booktitle}{\emph{Forty-first International Conference on Machine Learning, {ICML} 2024, Vienna, Austria, July 21-27, 2024}}. \bibinfo{publisher}{OpenReview.net}.
\newblock
\urldef\tempurl%
\url{https://openreview.net/forum?id=0NphYCmgua}
\showURL{%
\tempurl}


\bibitem[Zeng et~al\mbox{.}(2025)]%
        {autogen}
\bibfield{author}{\bibinfo{person}{Daojian Zeng}, \bibinfo{person}{Lin Zhou}, \bibinfo{person}{Zhiheng Zhang}, {and} \bibinfo{person}{Lincheng Jiang}.} \bibinfo{year}{2025}\natexlab{}.
\newblock \showarticletitle{AuToGen: Automated Tool Learning Data Generation with Domain-specific Structured Data}.
\newblock \bibinfo{journal}{\emph{Data Intell.}} \bibinfo{volume}{7}, \bibinfo{number}{4} (\bibinfo{year}{2025}), \bibinfo{pages}{1108--1128}.
\newblock
\href{https://doi.org/10.3724/2096-7004.DI.2024.0005}{doi:\nolinkurl{10.3724/2096-7004.DI.2024.0005}}


\bibitem[Zhang et~al\mbox{.}(2026a)]%
        {coevoskills}
\bibfield{author}{\bibinfo{person}{Hanrong Zhang}, \bibinfo{person}{Shichen Fan}, \bibinfo{person}{Henry~Peng Zou}, \bibinfo{person}{Yankai Chen}, \bibinfo{person}{Zhenting Wang}, \bibinfo{person}{Jiayuan Zhou}, \bibinfo{person}{Chengze Li}, \bibinfo{person}{Wei-Chieh Huang}, \bibinfo{person}{Yifei Yao}, \bibinfo{person}{Kening Zheng}, \bibinfo{person}{Xue Liu}, \bibinfo{person}{Xiaoxiao Li}, {and} \bibinfo{person}{Philip~S. Yu}.} \bibinfo{year}{2026}\natexlab{a}.
\newblock \showarticletitle{CoEvoSkills: Self-Evolving Agent Skills via Co-Evolutionary Verification}.
\newblock
\urldef\tempurl%
\url{https://api.semanticscholar.org/CorpusID:287071917}
\showURL{%
\tempurl}


\bibitem[Zhang et~al\mbox{.}(2026b)]%
        {funprm}
\bibfield{author}{\bibinfo{person}{Ruiyi Zhang}, \bibinfo{person}{Peijia Qin}, \bibinfo{person}{Qi Cao}, \bibinfo{person}{Eric Xue}, {and} \bibinfo{person}{Pengtao Xie}.} \bibinfo{year}{2026}\natexlab{b}.
\newblock \showarticletitle{FunPRM: Function-as-Step Process Reward Model with Meta Reward Correction for Code Generation}.
\newblock \bibinfo{journal}{\emph{arXiv preprint arXiv:2601.22249}} (\bibinfo{year}{2026}).
\newblock


\bibitem[Zhang et~al\mbox{.}(2025a)]%
        {deepanalyze}
\bibfield{author}{\bibinfo{person}{Shaolei Zhang}, \bibinfo{person}{Ju Fan}, \bibinfo{person}{Meihao Fan}, \bibinfo{person}{Guoliang Li}, {and} \bibinfo{person}{Xiaoyong Du}.} \bibinfo{year}{2025}\natexlab{a}.
\newblock \showarticletitle{DeepAnalyze: Agentic Large Language Models for Autonomous Data Science}.
\newblock \bibinfo{journal}{\emph{CoRR}}  \bibinfo{volume}{abs/2510.16872} (\bibinfo{year}{2025}).
\newblock
\showeprint[arXiv]{2510.16872}
\href{https://doi.org/10.48550/ARXIV.2510.16872}{doi:\nolinkurl{10.48550/ARXIV.2510.16872}}


\bibitem[Zhang et~al\mbox{.}(2025d)]%
        {self_rewarding_prm}
\bibfield{author}{\bibinfo{person}{Shimao Zhang}, \bibinfo{person}{Xiao Liu}, \bibinfo{person}{Xin Zhang}, \bibinfo{person}{Junxiao Liu}, \bibinfo{person}{Zheheng Luo}, \bibinfo{person}{Shujian Huang}, {and} \bibinfo{person}{Yeyun Gong}.} \bibinfo{year}{2025}\natexlab{d}.
\newblock \showarticletitle{Process-based Self-Rewarding Language Models}. In \bibinfo{booktitle}{\emph{Findings of the Association for Computational Linguistics, {ACL} 2025, Vienna, Austria, July 27 - August 1, 2025}}, \bibfield{editor}{\bibinfo{person}{Wanxiang Che}, \bibinfo{person}{Joyce Nabende}, \bibinfo{person}{Ekaterina Shutova}, {and} \bibinfo{person}{Mohammad~Taher Pilehvar}} (Eds.). \bibinfo{publisher}{Association for Computational Linguistics}, \bibinfo{pages}{18097--18110}.
\newblock
\urldef\tempurl%
\url{https://aclanthology.org/2025.findings-acl.930/}
\showURL{%
\tempurl}


\bibitem[Zhang et~al\mbox{.}(2025c)]%
        {dr_survey}
\bibfield{author}{\bibinfo{person}{Wenlin Zhang}, \bibinfo{person}{Xiaopeng Li}, \bibinfo{person}{Yingyi Zhang}, \bibinfo{person}{Pengyue Jia}, \bibinfo{person}{Yichao Wang}, \bibinfo{person}{Huifeng Guo}, \bibinfo{person}{Yong Liu}, {and} \bibinfo{person}{Xiangyu Zhao}.} \bibinfo{year}{2025}\natexlab{c}.
\newblock \showarticletitle{Deep Research: {A} Survey of Autonomous Research Agents}.
\newblock \bibinfo{journal}{\emph{CoRR}}  \bibinfo{volume}{abs/2508.12752} (\bibinfo{year}{2025}).
\newblock
\showeprint[arXiv]{2508.12752}
\href{https://doi.org/10.48550/ARXIV.2508.12752}{doi:\nolinkurl{10.48550/ARXIV.2508.12752}}


\bibitem[Zhang et~al\mbox{.}(2023)]%
        {datacopilot}
\bibfield{author}{\bibinfo{person}{Wenqi Zhang}, \bibinfo{person}{Yongliang Shen}, \bibinfo{person}{Weiming Lu}, {and} \bibinfo{person}{Yueting Zhuang}.} \bibinfo{year}{2023}\natexlab{}.
\newblock \showarticletitle{Data-Copilot: Bridging Billions of Data and Humans with Autonomous Workflow}.
\newblock \bibinfo{journal}{\emph{CoRR}}  \bibinfo{volume}{abs/2306.07209} (\bibinfo{year}{2023}).
\newblock
\showeprint[arXiv]{2306.07209}
\href{https://doi.org/10.48550/ARXIV.2306.07209}{doi:\nolinkurl{10.48550/ARXIV.2306.07209}}


\bibitem[Zhang et~al\mbox{.}(2024)]%
        {eff_tool_agent_da}
\bibfield{author}{\bibinfo{person}{Xilin Zhang}, \bibinfo{person}{Zhixin Mao}, \bibinfo{person}{Ziwen Chen}, {and} \bibinfo{person}{Shen Gao}.} \bibinfo{year}{2024}\natexlab{}.
\newblock \showarticletitle{Effective Tool Augmented Multi-Agent Framework for Data Analysis}.
\newblock \bibinfo{journal}{\emph{Data Intell.}} \bibinfo{volume}{6}, \bibinfo{number}{4} (\bibinfo{year}{2024}), \bibinfo{pages}{923--945}.
\newblock
\href{https://doi.org/10.3724/2096-7004.DI.2024.0013}{doi:\nolinkurl{10.3724/2096-7004.DI.2024.0013}}


\bibitem[Zhang et~al\mbox{.}(2025b)]%
        {reward-sql}
\bibfield{author}{\bibinfo{person}{Yuxin Zhang}, \bibinfo{person}{Meihao Fan}, \bibinfo{person}{Ju Fan}, \bibinfo{person}{Mingyang Yi}, \bibinfo{person}{Yuyu Luo}, \bibinfo{person}{Jian Tan}, {and} \bibinfo{person}{Guoliang Li}.} \bibinfo{year}{2025}\natexlab{b}.
\newblock \showarticletitle{Reward-SQL: Boosting Text-to-SQL via Stepwise Reasoning and Process-Supervised Rewards}.
\newblock \bibinfo{journal}{\emph{CoRR}}  \bibinfo{volume}{abs/2505.04671} (\bibinfo{year}{2025}).
\newblock
\showeprint[arXiv]{2505.04671}
\href{https://doi.org/10.48550/ARXIV.2505.04671}{doi:\nolinkurl{10.48550/ARXIV.2505.04671}}


\bibitem[Zhang et~al\mbox{.}(2025e)]%
        {qwenprm}
\bibfield{author}{\bibinfo{person}{Zhenru Zhang}, \bibinfo{person}{Chujie Zheng}, \bibinfo{person}{Yangzhen Wu}, \bibinfo{person}{Beichen Zhang}, \bibinfo{person}{Runji Lin}, \bibinfo{person}{Bowen Yu}, \bibinfo{person}{Dayiheng Liu}, \bibinfo{person}{Jingren Zhou}, {and} \bibinfo{person}{Junyang Lin}.} \bibinfo{year}{2025}\natexlab{e}.
\newblock \showarticletitle{The Lessons of Developing Process Reward Models in Mathematical Reasoning}. In \bibinfo{booktitle}{\emph{Findings of the Association for Computational Linguistics, {ACL} 2025, Vienna, Austria, July 27 - August 1, 2025}}, \bibfield{editor}{\bibinfo{person}{Wanxiang Che}, \bibinfo{person}{Joyce Nabende}, \bibinfo{person}{Ekaterina Shutova}, {and} \bibinfo{person}{Mohammad~Taher Pilehvar}} (Eds.). \bibinfo{publisher}{Association for Computational Linguistics}, \bibinfo{pages}{10495--10516}.
\newblock
\urldef\tempurl%
\url{https://aclanthology.org/2025.findings-acl.547/}
\showURL{%
\tempurl}


\bibitem[Zhao et~al\mbox{.}(2025b)]%
        {genprm}
\bibfield{author}{\bibinfo{person}{Jian Zhao}, \bibinfo{person}{Runze Liu}, \bibinfo{person}{Kaiyan Zhang}, \bibinfo{person}{Zhimu Zhou}, \bibinfo{person}{Junqi Gao}, \bibinfo{person}{Dong Li}, \bibinfo{person}{Jiafei Lyu}, \bibinfo{person}{Zhouyi Qian}, \bibinfo{person}{Biqing Qi}, \bibinfo{person}{Xiu Li}, {and} \bibinfo{person}{Bowen Zhou}.} \bibinfo{year}{2025}\natexlab{b}.
\newblock \showarticletitle{GenPRM: Scaling Test-Time Compute of Process Reward Models via Generative Reasoning}.
\newblock \bibinfo{journal}{\emph{CoRR}}  \bibinfo{volume}{abs/2504.00891} (\bibinfo{year}{2025}).
\newblock
\showeprint[arXiv]{2504.00891}
\href{https://doi.org/10.48550/ARXIV.2504.00891}{doi:\nolinkurl{10.48550/ARXIV.2504.00891}}


\bibitem[Zhao et~al\mbox{.}(2025a)]%
        {ms_swift}
\bibfield{author}{\bibinfo{person}{Yuze Zhao}, \bibinfo{person}{Jintao Huang}, \bibinfo{person}{Jinghan Hu}, \bibinfo{person}{Xingjun Wang}, \bibinfo{person}{Yunlin Mao}, \bibinfo{person}{Daoze Zhang}, \bibinfo{person}{Zeyinzi Jiang}, \bibinfo{person}{Zhikai Wu}, \bibinfo{person}{Baole Ai}, \bibinfo{person}{Ang Wang}, \bibinfo{person}{Wenmeng Zhou}, {and} \bibinfo{person}{Yingda Chen}.} \bibinfo{year}{2025}\natexlab{a}.
\newblock \showarticletitle{{SWIFT:} {A} Scalable Lightweight Infrastructure for Fine-Tuning}. In \bibinfo{booktitle}{\emph{AAAI-25, Sponsored by the Association for the Advancement of Artificial Intelligence, February 25 - March 4, 2025, Philadelphia, PA, {USA}}}, \bibfield{editor}{\bibinfo{person}{Toby Walsh}, \bibinfo{person}{Julie Shah}, {and} \bibinfo{person}{Zico Kolter}} (Eds.). \bibinfo{publisher}{{AAAI} Press}, \bibinfo{pages}{29733--29735}.
\newblock
\href{https://doi.org/10.1609/AAAI.V39I28.35383}{doi:\nolinkurl{10.1609/AAAI.V39I28.35383}}


\bibitem[Zheng et~al\mbox{.}(2025)]%
        {prmsurvey}
\bibfield{author}{\bibinfo{person}{Congming Zheng}, \bibinfo{person}{Jiachen Zhu}, \bibinfo{person}{Zhuoying Ou}, \bibinfo{person}{Yuxiang Chen}, \bibinfo{person}{Kangning Zhang}, \bibinfo{person}{Rong Shan}, \bibinfo{person}{Zeyu Zheng}, \bibinfo{person}{Mengyue Yang}, \bibinfo{person}{Jianghao Lin}, \bibinfo{person}{Yong Yu}, {and} \bibinfo{person}{Weinan Zhang}.} \bibinfo{year}{2025}\natexlab{}.
\newblock \showarticletitle{A Survey of Process Reward Models: From Outcome Signals to Process Supervisions for Large Language Models}.
\newblock \bibinfo{journal}{\emph{CoRR}}  \bibinfo{volume}{abs/2510.08049} (\bibinfo{year}{2025}).
\newblock
\showeprint[arXiv]{2510.08049}
\href{https://doi.org/10.48550/ARXIV.2510.08049}{doi:\nolinkurl{10.48550/ARXIV.2510.08049}}


\bibitem[Zheng et~al\mbox{.}(2023)]%
        {llm_as_judge}
\bibfield{author}{\bibinfo{person}{Lianmin Zheng}, \bibinfo{person}{Wei{-}Lin Chiang}, \bibinfo{person}{Ying Sheng}, \bibinfo{person}{Siyuan Zhuang}, \bibinfo{person}{Zhanghao Wu}, \bibinfo{person}{Yonghao Zhuang}, \bibinfo{person}{Zi Lin}, \bibinfo{person}{Zhuohan Li}, \bibinfo{person}{Dacheng Li}, \bibinfo{person}{Eric~P. Xing}, \bibinfo{person}{Hao Zhang}, \bibinfo{person}{Joseph~E. Gonzalez}, {and} \bibinfo{person}{Ion Stoica}.} \bibinfo{year}{2023}\natexlab{}.
\newblock \showarticletitle{Judging LLM-as-a-Judge with MT-Bench and Chatbot Arena}. In \bibinfo{booktitle}{\emph{Advances in Neural Information Processing Systems 36: Annual Conference on Neural Information Processing Systems 2023, NeurIPS 2023, New Orleans, LA, USA, December 10 - 16, 2023}}, \bibfield{editor}{\bibinfo{person}{Alice Oh}, \bibinfo{person}{Tristan Naumann}, \bibinfo{person}{Amir Globerson}, \bibinfo{person}{Kate Saenko}, \bibinfo{person}{Moritz Hardt}, {and} \bibinfo{person}{Sergey Levine}} (Eds.).
\newblock
\urldef\tempurl%
\url{http://papers.nips.cc/paper\_files/paper/2023/hash/91f18a1287b398d378ef22505bf41832-Abstract-Datasets\_and\_Benchmarks.html}
\showURL{%
\tempurl}


\bibitem[Zhou et~al\mbox{.}(2025)]%
        {finprm}
\bibfield{author}{\bibinfo{person}{Yuanchen Zhou}, \bibinfo{person}{Shuo Jiang}, \bibinfo{person}{Jie Zhu}, \bibinfo{person}{Junhui Li}, \bibinfo{person}{Lifan Guo}, \bibinfo{person}{Feng Chen}, {and} \bibinfo{person}{Chi Zhang}.} \bibinfo{year}{2025}\natexlab{}.
\newblock \showarticletitle{Fin-PRM: {A} Domain-Specialized Process Reward Model for Financial Reasoning in Large Language Models}.
\newblock \bibinfo{journal}{\emph{CoRR}}  \bibinfo{volume}{abs/2508.15202} (\bibinfo{year}{2025}).
\newblock
\showeprint[arXiv]{2508.15202}
\href{https://doi.org/10.48550/ARXIV.2508.15202}{doi:\nolinkurl{10.48550/ARXIV.2508.15202}}


\bibitem[Zhu et~al\mbox{.}(2025a)]%
        {dataagent_survey}
\bibfield{author}{\bibinfo{person}{Yizhang Zhu}, \bibinfo{person}{Liangwei Wang}, \bibinfo{person}{Chenyu Yang}, \bibinfo{person}{Xiaotian Lin}, \bibinfo{person}{Boyan Li}, \bibinfo{person}{Wei Zhou}, \bibinfo{person}{Xinyu Liu}, \bibinfo{person}{Zhangyang Peng}, \bibinfo{person}{Tianqi Luo}, \bibinfo{person}{Yu Li}, \bibinfo{person}{Chengliang Chai}, \bibinfo{person}{Chong Chen}, \bibinfo{person}{Shimin Di}, \bibinfo{person}{Ju Fan}, \bibinfo{person}{Ji Sun}, \bibinfo{person}{Nan Tang}, \bibinfo{person}{Fugee Tsung}, \bibinfo{person}{Jiannan Wang}, \bibinfo{person}{Chenglin Wu}, \bibinfo{person}{Yanwei Xu}, \bibinfo{person}{Shaolei Zhang}, \bibinfo{person}{Yong Zhang}, \bibinfo{person}{Xuanhe Zhou}, \bibinfo{person}{Guoliang Li}, {and} \bibinfo{person}{Yuyu Luo}.} \bibinfo{year}{2025}\natexlab{a}.
\newblock \showarticletitle{A Survey of Data Agents: Emerging Paradigm or Overstated Hype?}
\newblock \bibinfo{journal}{\emph{ArXiv}}  \bibinfo{volume}{abs/2510.23587} (\bibinfo{year}{2025}).
\newblock
\urldef\tempurl%
\url{https://api.semanticscholar.org/CorpusID:282389107}
\showURL{%
\tempurl}


\bibitem[Zhu et~al\mbox{.}(2025b)]%
        {dataanalysis-study}
\bibfield{author}{\bibinfo{person}{Yuqi Zhu}, \bibinfo{person}{Yi Zhong}, \bibinfo{person}{Jintian Zhang}, \bibinfo{person}{Ziheng Zhang}, \bibinfo{person}{Shuofei Qiao}, \bibinfo{person}{Yujie Luo}, \bibinfo{person}{Lun Du}, \bibinfo{person}{Da Zheng}, \bibinfo{person}{Huajun Chen}, {and} \bibinfo{person}{Ningyu Zhang}.} \bibinfo{year}{2025}\natexlab{b}.
\newblock \showarticletitle{Why Do Open-Source LLMs Struggle with Data Analysis? {A} Systematic Empirical Study}.
\newblock \bibinfo{journal}{\emph{CoRR}}  \bibinfo{volume}{abs/2506.19794} (\bibinfo{year}{2025}).
\newblock
\showeprint[arXiv]{2506.19794}
\href{https://doi.org/10.48550/ARXIV.2506.19794}{doi:\nolinkurl{10.48550/ARXIV.2506.19794}}


\bibitem[Zou et~al\mbox{.}(2025a)]%
        {tattoo}
\bibfield{author}{\bibinfo{person}{Jiaru Zou}, \bibinfo{person}{Soumya Roy}, \bibinfo{person}{Vinay~Kumar Verma}, \bibinfo{person}{Ziyi Wang}, \bibinfo{person}{David Wipf}, \bibinfo{person}{Pan Lu}, \bibinfo{person}{Sumit Negi}, \bibinfo{person}{James Zou}, {and} \bibinfo{person}{Jingrui He}.} \bibinfo{year}{2025}\natexlab{a}.
\newblock \showarticletitle{TaTToo: Tool-Grounded Thinking {PRM} for Test-Time Scaling in Tabular Reasoning}.
\newblock \bibinfo{journal}{\emph{CoRR}}  \bibinfo{volume}{abs/2510.06217} (\bibinfo{year}{2025}).
\newblock
\showeprint[arXiv]{2510.06217}
\href{https://doi.org/10.48550/ARXIV.2510.06217}{doi:\nolinkurl{10.48550/ARXIV.2510.06217}}


\bibitem[Zou et~al\mbox{.}(2025b)]%
        {reasonflux-prm}
\bibfield{author}{\bibinfo{person}{Jiaru Zou}, \bibinfo{person}{Ling Yang}, \bibinfo{person}{Jingwen Gu}, \bibinfo{person}{Jiahao Qiu}, \bibinfo{person}{Ke Shen}, \bibinfo{person}{Jingrui He}, {and} \bibinfo{person}{Mengdi Wang}.} \bibinfo{year}{2025}\natexlab{b}.
\newblock \showarticletitle{ReasonFlux-PRM: Trajectory-Aware PRMs for Long Chain-of-Thought Reasoning in LLMs}.
\newblock \bibinfo{journal}{\emph{CoRR}}  \bibinfo{volume}{abs/2506.18896} (\bibinfo{year}{2025}).
\newblock
\showeprint[arXiv]{2506.18896}
\href{https://doi.org/10.48550/ARXIV.2506.18896}{doi:\nolinkurl{10.48550/ARXIV.2506.18896}}


\end{thebibliography}

\appendix

\section{Theoretical Perspective for Environment-Aware Verifier}
\label{app:theoretical_perspective}
We formalize data analysis as a Partially Observable Markov Decision Process (POMDP) where the true environment state is a latent variable $\varepsilon$.
To evaluate an agent's trajectory, traditional static PRMs must implicitly estimate this unknown environment by relying on an internal prior distribution $P_{\mathrm{prior}}(\varepsilon| h_{t})$ learned during training. However, real-world scientific data is highly heterogeneous and frequently out-of-distribution ($\varepsilon_{\mathrm{true}} \notin P_{\mathrm{prior}}$).
This uncertainty causes the "Incorrect Rewarding for Silent Errors" (Tab.\ref{tab:errors}), where the PRM hallucinates a compatible environment.

DataPRM mitigates this via explicit interaction, drawing ground-truth observations $o_t \sim P(O \mid \varepsilon, a_t, h_t)$ to update the uncertain prior into an accurate posterior via Bayes' theorem:
$$
P_{\mathrm{post}}(\varepsilon \mid o_t, a_t, h_t) \propto P(o_t \mid \varepsilon, a_t) \cdot P_{\mathrm{prior}}(\varepsilon \mid h_{t})
$$

Mechanistically, environmental interaction acts as a necessary Bayesian evidence-gathering step that grounds latent variables and reduces reward estimator variance.

Furthermore, this Bayesian perspective rigorously derives our ternary reward. In an exploratory POMDP, optimal steps require balancing exploitation (task progress) and exploration (uncertainty reduction). Therefore, the reward of an agent's step $R(a_t)$ can be theoretically composed of two parts Progress towards the Final Goal $G$ and Information Gain about the Hidden Environment $I$. We formalize the reward as a balanced combination ($\lambda$=0.5):

$$
R(a_t)= \lambda \cdot G(a_t)+ (1-\lambda) \cdot I(a_t)
$$

Here, $I(a_t)$ is the KL divergence $D_{\mathrm{KL}}(P_{\mathrm{post}} \| P_{\mathrm{prior}})$. Because reliably annotating continuous rewards for KL divergence is intractable in practice, we approximate the Information Gain using an indicator function, $\mathbb{I}[I(a_t)>\epsilon]$ for a small threshold $\epsilon$, signifying effective information gain. This maps exactly to our 3-value mechanism:

\begin{itemize}[leftmargin=*]
    \item Strictly Correct ($R=1$): The action makes progress on the task ($G=1$) and confirms the validity of the current logic ($\mathbb{I}=1$).
    \item Grounding / Correctable Error ($R=0.5$): The action fails to make direct task progress ($G=0$), but the resulting observation provides critical information about the environment ($\mathbb{I}=1$).
    \item Irrecoverable Error ($R=0$): The action makes no progress ($G=0$) and yields no information or produces hallucinations ($\mathbb{I}=0$).
\end{itemize}

\end{document}